\documentclass{article}

\usepackage{amsmath,amsfonts,bm}

\newcommand{\TV}{\mathrm{TV}}

\def\eqref#1{equation~\ref{#1}}

\def\1{\bm{1}}

\def\tr{\operatorname{tr}}

\DeclareMathAlphabet{\mathsfit}{\encodingdefault}{\sfdefault}{m}{sl}
\SetMathAlphabet{\mathsfit}{bold}{\encodingdefault}{\sfdefault}{bx}{n}

\newcommand{\E}{\mathbb{E}}

\newcommand{\norm}[1]{\left\lVert#1\right\rVert}

\PassOptionsToPackage{sort&compress}{natbib}
\usepackage[preprint]{neurips_2026}

\makeatletter
\renewcommand{\@notice}{}
\makeatother

\usepackage[utf8]{inputenc} %
\usepackage[T1]{fontenc}    %
\usepackage{amsthm}
\usepackage{thmtools}
\usepackage{amssymb,mathtools}
\usepackage{bm}
\usepackage{graphicx}
\usepackage{booktabs}
\usepackage{array}
\usepackage{makecell}
\usepackage{tabularx}
\usepackage{enumitem}
\usepackage{nicefrac}
\usepackage{algorithm}
\usepackage{algpseudocode}
\usepackage{float}
\usepackage{url}
\usepackage{hyperref}
\usepackage{wrapfig}
\usepackage[createShortEnv,]{proof-at-the-end}
\usepackage[section, above]{placeins}
\usepackage{siunitx}
\usepackage{multirow}
\usepackage{multicol}

\usepackage[capitalize,noabbrev,nameinlink]{cleveref}

\AddToHook{cmd/appendix/before}{%
    \crefalias{section}{appendix}%
    \crefalias{subsection}{appendix}
}
\usepackage[final]{microtype}      %
\usepackage[dvipsnames, x11names]{xcolor}         %

\usepackage{tikz}
\usepackage{subcaption}
\usetikzlibrary{automata, positioning, arrows.meta,fit,backgrounds,calc,shapes.geometric}
\usepackage{changepage}

\theoremstyle{plain}

\newcommand{\method}{\textsc{MA-JEPA}}

\newcommand{\res}[2]{#1\,{\scriptsize (#2)}}
\newcommand{\best}[2]{\textbf{#1}\,{\normalfont\scriptsize (#2)}}

\title{MA-JEPA: Joint-Embedding World Models for Multi-Agent Reinforcement Learning}

\author{%
  Brandon  Gary Kaplowitz\textsuperscript{*} \quad
  Osaze James Obahor\textsuperscript{*} \quad
  Christian Schroeder de Witt \\
  Department of Engineering Science, University of Oxford \\
  \textsuperscript{*}Equal contribution. \quad Correspondence to \texttt{brandon.kaplowitz@eng.ox.ac.uk}.
}
\begin{document}

\maketitle
\begin{abstract}
World models can improve sample efficiency in reinforcement learning by training policies on imagined trajectories, but their effectiveness depends on learning representations that preserve the information needed for prediction and control. We introduce \method{}, a joint-embedding predictive world model for multi-agent reinforcement learning under centralized training and decentralized execution. Rather than reconstructing observations, \method{} learns latent dynamics by predicting future observation embeddings. During centralized training, a joint predictor uses all agents' local states and actions to predict each agent's next observation embedding, which is combined with that agent's independently maintained local dynamics to construct the next state used by its policy. On eight StarCraft Multi-Agent Challenge tasks, we match or exceeds the strongest model-based comparator mean win rate on four maps and outperforms the strongest model-free baseline on seven of eight maps, demonstrating that joint-embedding prediction can support effective multi-agent control without observation reconstruction.
\end{abstract}

\section{Introduction}
\label{sec:introduction}
Model-based reinforcement learning (MBRL) learns an environment model and uses it to generate trajectories for policy improvement.  By training an actor and critic on these imagined trajectories, methods can reuse each real transition for many learning updates \citep{hafner2019planet,hafner2019dream,hafner2023dreamerv3}. The benefit, however, depends on the quality of the world model.
Errors can accumulate as the model is rolled forward, and a policy trained inside the model may exploit predictions that do not match the real environment \citep{janner2019mbpo,moerland2023mbrl}.

Multi-agent MBRL adds a second problem, where an agent's transition probabilities generally depend on the actions of the other agents.
For example, in the cooperative combat environments of the StarCraft Multi-Agent Challenge (SMAC), whether an opponent remains alive after one step depends on which target the whole team attacks. 
A local world model conditioned only on one agent's action cannot fully predict this outcome. 
A centralized model can condition on the joint action, but directly using a centralized latent state for policy learning leads to a different problem, where the actor is then trained on information that is unavailable during decentralized execution. 
Existing multi-agent model-based reinforcement learning does not fully address the problem \citep{egorov2022mamba,venugopal2024mabl,zhang2024marie,deihim2025matwm}.

In this paper, we directly tackle this problem by introducing MA-JEPA, the first joint-embedding model-based multi-agent reinforcement learning (MARL) method. It achieves centralized training with decentralized execution (CTDE) by training on the same latent representation of observations in centralized training that it does in decentralized execution. \citep{oliehoek2008,kraemer2016} 
Our agents explicitly maintain an independent recurrent state in a parameter-shared local world model. This enables us both to train agents using local latent states and to maintain parallel, independent copies of the world model during execution, each with information specific to an individual agent. 

We train this model with a joint-embedding predictive architecture (JEPA) objective that predicts target observation embeddings from latent states and action histories, without reconstructing
observations, unlike leading model-based MARL methods \citep{assran2023ijepa,bardes2024vjepa,assran2025vjepa2}.
During multi-agent training, a joint predictor conditions on all agents'
local states and actions to predict an embedding of each agent's
next local observation. During real interactions, the predicted embedding is combined with the agent's local dynamics state to construct its next local state for policy learning. During centralized imagination, it combines the local dynamics state with the embedding predicted from the joint action. The actor, therefore, receives the same type of locally constructed state in both cases. A centralized critic uses all local states for value learning. Importantly, neither it nor the joint predictor runs during execution, meaning only agent-specific information is used during execution.

We evaluate the full method on SMAC at matched interaction budgets and compare the effects of outcome and embedding gradients on the local representations.

Our contributions include:
\begin{enumerate}[leftmargin=*,itemsep=2pt,topsep=3pt]
\item A model-based architecture for CTDE that predicts each agent's next local observation embedding from all agents' local states and actions, then uses the same local posterior as during execution to construct the actor state.

\item To our knowledge, the first JEPA-based MARL approach, which trains local and joint dynamics by predicting target
embeddings instead of reconstructing observations, enabling the same
representation-learning approach in single-agent and multi-agent
imagination.
\end{enumerate}

\section{Preliminaries}
\label{sec:preliminaries}

\subsection{Model-Based Multi-Agent Reinforcement Learning}
\label{sec:marl-background}

We consider a cooperative decentralized partially observable Markov decision
process (Dec-POMDP) with $N$ agents. At timestep $t$, the environment is in
state $x_t\in\mathcal X$, agent $i$ receives a local observation
$\mathbf o_t^i\in\mathcal O^i$, and selects an action
$\mathbf a_t^i\in\mathcal A^i$. The individual actions form the joint action
\begin{equation}
\mathbf a_t^{1:N}
=
(\mathbf a_t^1,\ldots,\mathbf a_t^N),
\end{equation}
after which the environment transitions according to
$p(x_{t+1}\mid x_t,\mathbf a_t^{1:N})$ and produces a reward $r_{t+1}$.

Because the environment state is not available during decentralized
execution, agent $i$ acts from its local interaction history
\begin{equation}
\tau_t^i
=
(\mathbf o_1^i,\mathbf a_1^i,\ldots,\mathbf o_t^i),
\qquad
\mathbf a_t^i
\sim
\pi^i(\cdot\mid\tau_t^i).
\label{eq:decentralized-policy}
\end{equation}
Under CTDE, information from all agents may be used to learn models and value functions,
while the executed policy remains conditioned only on information available
in $\tau_t^i$.

Throughout the paper, superscript $i$ indexes agents and
$\mathbf x_t^{1:N}=(\mathbf x_t^1,\ldots,\mathbf x_t^N)$ denotes a
synchronized collection of agent-indexed quantities. We use hats for quantities
generated by a learned model and $\operatorname{sg}(\cdot)$ for
stop-gradient.

Model-based reinforcement learning learns environment dynamics from collected
experience and uses the learned model to generate trajectories for policy
improvement \citep{moerland2023mbrl}. In the multi-agent setting, imagined
rollouts are generated from replay-inferred states while actions are sampled
from the decentralized policies, and the model predicts the quantities required to advance the rollout,
including rewards and continuation factors
$\widehat r_{t+1}^i$ and $\widehat c_{t+1}^i$.
We define $\widehat c_{t+1}^i$ as the discounted continuation factor, i.e.,
it incorporates both discounting and episode termination.

A critic may use centralized information during training. Let
$V^i(\widehat{\mathbf s}_t^{1:N})$ denote the value estimate for agent $i$
from the synchronized imagined states. Imagined $\lambda$-returns are
\begin{equation}
G_t^{\lambda,i}
=
\widehat r_{t+1}^i
+\widehat c_{t+1}^i
\left[
(1-\lambda)
V^i(\widehat{\mathbf s}_{t+1}^{1:N})
+
\lambda G_{t+1}^{\lambda,i}
\right],
\label{eq:prelim-lambda-return}
\end{equation}
where $\lambda\in[0,1]$ controls the trade-off between value bootstrapping and
longer imagined returns \citep{sutton1988td,hafner2023dreamerv3}. The actor is optimized from imagined returns while centralized information remains
restricted to training.

\subsection{Joint-Embedding Predictive Architectures}
\label{sec:jepa-background}

Joint-embedding predictive architectures (JEPAs) learn representations by
predicting the embedding of a target from a related context instead of
reconstructing the target itself
\citep{lecun2022path,assran2023ijepa}. Let $x$ denote a context and $y$ a
prediction target. An online encoder $E_\phi$, a target encoder
$E_{\bar\phi}$, and a predictor $P_\theta$ produce
\begin{equation}
\mathbf e_x
=
E_\phi(x),
\qquad
\bar{\mathbf e}_y
=
\operatorname{sg}\!\left(E_{\bar\phi}(y)\right),
\qquad
\widehat{\mathbf e}_y
=
P_\theta(\mathbf e_x),
\label{eq:jepa-representations}
\end{equation}
with objective
\begin{equation}
\mathcal L_{\mathrm{JEPA}}
=
D\!\left(
\widehat{\mathbf e}_y,
\bar{\mathbf e}_y
\right),
\label{eq:standard-jepa}
\end{equation}
where $D$ measures discrepancy in representation space. The target encoder is
typically updated as an exponential moving average (EMA) of the online encoder
\citep{grill2020byol,assran2023ijepa}. Since predictive agreement can admit
collapsed representations, practical JEPA methods additionally employ
architectural asymmetry or representation regularization; we use SIGReg
\citep{balestriero2025lejepa}.

For sequential decision making, prediction is additionally conditioned on the
actions connecting the current context to a future observation. At horizon
$h$, this takes the general form
\begin{equation}
\widehat{\mathbf e}_{t+h}
=
P_\theta\!\left(
\mathbf c_t,
\mathbf a_{t:t+h-1}
\right),
\qquad
\bar{\mathbf e}_{t+h}
=
\operatorname{sg}\!\left(
E_{\bar\phi}(\mathbf o_{t+h})
\right),
\label{eq:action-conditioned-jepa}
\end{equation}
where $\mathbf c_t$ denotes the available predictive context.
Action-conditioned joint-embedding models have been used for latent prediction,
planning, and control
\citep{assran2025vjepa2,maes2026leworldmodel}.

For the remainder of the paper,
$\mathbf e_t^i=E_\phi(\mathbf o_t^i)$ denotes the online observation embedding
of agent $i$, $\bar{\mathbf e}_t^i$ its stopped EMA target, and
$\widehat{\mathbf e}_t^i$ a predicted observation embedding. We reserve
$\mathbf z_t^i$ for the stochastic latent state inferred by the local world
model. In \method{} the joint model predicts
observation embeddings, which are subsequently processed by the local
world-model posterior to construct the latent state used by the policy.

\section{Related Work}
\label{sec:related-work}

\subsection{Joint-Embedding Predictive Architectures}
\label{sec:related-jepa}

Joint-embedding predictive architectures learn representations by predicting
target embeddings rather than reconstructing observations
\citep{lecun2022path}. I-JEPA applies this principle to spatially masked image
regions, demonstrating that predictive objectives can learn semantic visual
representations without pixel-level reconstruction
\citep{assran2023ijepa}. V-JEPA extends the approach to video by predicting
masked spatiotemporal representations, while V-JEPA~2 scales video
pretraining and introduces an action-conditioned predictor for physical
reasoning and planning
\citep{bardes2024vjepa,assran2025vjepa2}. These works establish joint-embedding prediction as an effective alternative to reconstructive generative modeling for learning structured representations of visual dynamics.

Recent work has brought this principle closer to reinforcement learning and
world modeling. LeWorldModel (LeWM) trains an action-conditioned JEPA
end-to-end from pixels and uses SIGReg to stabilize the learned latent space,
showing that future representation prediction can support efficient
model-based control without an observation decoder
\citep{maes2026leworldmodel}. TD-JEPA instead develops policy-conditioned
long-horizon latent prediction for zero-shot reinforcement learning, using
temporal-difference structure to learn representations of future dynamics
across policies \citep{bagatella2025tdjepa}. Existing applications have
primarily focused on single-agent representation learning, planning, and
control. \method{} extends this architectural family to MARL,
using the learned embedding space both as the predictive target of each local
world model and as the interface through which a centralized joint model
captures multi-agent interactions.

\subsection{Model-Based Reinforcement Learning}
\label{sec:related-mbrl}

Modern model-based reinforcement learning learns compact world models and uses
their predictions for planning or policy optimization. PlaNet introduced
stochastic latent dynamics for planning from high-dimensional observations,
and the Dreamer family instead trains actor-critic policies directly on latent
imagined trajectories
\citep{hafner2019planet,hafner2019dream,hafner2023dreamerv3}. More recent
world models increasingly use sequence-modeling architectures. IRIS represents
visual observations with discrete tokens and models their dynamics
autoregressively with a Transformer, while STORM combines stochastic latent
states with Transformer dynamics for efficient imagination
\citep{micheli2023iris,zhang2023storm}. 

Extending world models to MARL additionally requires modeling interactions
between agents while preserving decentralized policies. Existing methods have largely adapted established single-agent world-model architectures to this
setting. MAMBA extends Dreamer-style latent imagination to cooperative MARL,
using parameter-shared local world models together with inter-agent
communication \citep{egorov2022mamba}. DMAWM similarly builds on DreamerV2, separating local agent modules from a shared environment model and aligning imagined latent-state distributions
with those inferred from local observations \citep{Xue2026}. In contrast, MATWM develops
a multi-agent extension of the STORM architecture, combining stochastic Transformer dynamics with teammate prediction, decentralized imagination, and a semi-centralized critic \citep{deihim2025matwm}. MARIE also adopts Transformer-based world modeling, coupling decentralized agent dynamics with a
centralized Perceiver that aggregates information across agents
\citep{zhang2024marie}. MABL instead introduces global and local latent
variables so that centralized information can improve world-model learning
while policies remain decentralized \citep{venugopal2024mabl}.

\method{} follows a different architectural lineage. Rather than adapting a
Dreamer- or STORM-style world model to the multi-agent setting, we are the first paper, to our knowledge, to leverage
joint-embedding predictive world modeling in a MARL setting. 
\method{} combines joint-embedding predictive learning with centralized
imagination and decentralized latent-state based policies for execution. Unlike MARIE, which trains policies on reconstructed observations, our agents act on latent
states maintained by their local world models. These states summarize
each agent's observation-action history, allowing its policy to use the
temporal representations learned through dynamics prediction. Both local and joint models learn by predicting observation representations rather than reconstructing observations, avoiding the need to preserve details that are only useful for reconstruction.

During training, a separate joint model captures interactions using all agents' local states and joint actions. DMAWM generates imagined latent states through a shared environment model and aligns them with states inferred locally. In \method{}, the joint model instead supplies predictions to each agent's local world model, which constructs the agent's next state in the same way as during real interaction. The actor is therefore trained on states produced by the same local world model it uses at execution, without requiring centralized information at test time.

\section{Theoretical Background}
\label{sec:theory}
We study identification and transition estimation in an action-conditioned Gaussian model under assumptions similar to \citet{Klindt2026}. \Cref{app:additional-theory} gives consistency results for empirical successor distributions and population optima of unrestricted autoregressive models under fixed representations and collection policies.
\paragraph{Action-conditioned Gaussian identification.}
Following the notation of \citet{Klindt2026}, let $z,z'\in\mathbb R^n$
be consecutive physical states. Let $x=g(z)$ be the joint observation and
$y=f(x)=h(z)\in\mathbb R^n$ the learned representation, where $f$ and $g$
are measurable. Assume $z\sim\mathcal N(0,I_n)$ and let $A$ be a joint action
in a finite set $\mathcal A$, independent of $z$. For each action
$a\in\mathcal A$, let $B_a\in\mathbb R^{n\times n}$ and assume
\begin{equation*}
 z'\mid z,A=a\sim\mathcal N(B_az,I_n-B_aB_a^\top),\qquad
 \|B_a\|_{\mathrm{op}}\leq1.
\end{equation*}
Define $\kappa_B=\lambda_{\min}(\mathbb E_A[B_AB_A^\top])
-\mathbb E_A\|B_A\|_{\mathrm{op}}^4$ and assume $\kappa_B>0$.
Let $J_\omega(y,a)$ predict the next representation. Consider minimizing
the population loss
$\mathcal L=\mathbb E\|h(z')-J_\omega(h(z),A)\|^2$
over encoder and predictor classes, subject to
$\mathbb{E}\left[h(z)\right]=0$ and $\operatorname{Cov}(h(z))=I_n$.
Assume these classes contain the encoder $h_0(z)=Q_0z$ and predictor
$J_{\omega_0}(y,a)=Q_0B_aQ_0^\top y$ for some orthogonal matrix $Q_0$.

For each observed action $a$, define $\widehat C_{m,a}$ as the empirical
mean of $y'y^\top$ over samples with action $a$. Define $\widehat V_{m,a}$
as the empirical mean of
$(y'-\widehat C_{m,a}y)(y'-\widehat C_{m,a}y)^\top$
over those same samples. For each observed action, set
$\widehat P_m(\cdot\mid y,a)=\mathcal N(\widehat C_{m,a}y,\widehat V_{m,a})$.
For an action not yet observed, define $\widehat P_m(\cdot\mid y,a)$ to
be any fixed Gaussian distribution.

\begin{theoremE}[Identification and transition consistency][end, restate, text link section]
\label{thm:gaussian-identifiability-summary}
The minimum of the population loss $\mathcal L$ over the admissible
encoder--predictor pairs is
$n-\operatorname{tr}\mathbb E_A[B_AB_A^\top]$.
For every globally minimizing pair, there is an orthogonal matrix $Q$
such that its encoder satisfies $h(z)=Qz$ almost surely.
For any such encoder, let $y'=h(z')$. The conditional distribution of $y'$ given $y=h(z)$ and $A=a$ admits the Gaussian version
\[P(\cdot\mid y,a)=\mathcal{N}(QB_aQ^\top y,I_n-QB_aB_a^\top Q^\top).\]
Hold this globally minimizing encoder fixed and collect $m$ fresh independent and identically distributed triples $(y,A,y')$ from the joint distribution induced by the assumed model for $(z,A,z')$ and this fixed encoder.

As $m\to\infty$, $\widehat P_m(\cdot\mid y,a)$ converges weakly to $P(\cdot\mid y,a)$ almost surely for each fixed $y\in\mathbb{R}^n$ and
every action $a\in\mathcal A$ with $\Pr(A=a)>0$.
For each such action satisfying $\|B_a\|_{\mathrm{op}}<1$, this convergence also holds in total variation.
\end{theoremE}
\begin{textAtEnd}
This extends the Hermite argument of \citet{Klindt2026} to joint-action prediction.
The full covariance allows cross-agent dependence. The result recovers
population optima of a modified objective followed by estimation with a
fixed encoder.
\end{textAtEnd}
\begin{proofE}
\emph{Identification.} Decompose the centered encoder into its linear and
higher Hermite degrees:
\[
 h(z)=Mz+\nu(z),\qquad M=\E[h(z)z^\top],\qquad
 W=\E\norm{\nu(z)}^2.
\]
Whitening gives $MM^\top+\E[\nu\nu^\top]=I_n$, hence
$I_n-M^\top M\succeq0$ and $\tr(I_n-M^\top M)=W$.

Let $(T_av)(z)=\E[v(z')\mid z,A=a]$.
Under the singular value decomposition
$B_a=U\operatorname{diag}(\sigma_j)V^\top$, the rotated coordinates
$V^\top z,U^\top z'$ form independent Gaussian pairs with correlations
$\sigma_j$. The Hermite generating function maps normalized products of
coordinate degrees $\alpha_j$ between these rotated bases with multipliers
$\prod_j\sigma_j^{\alpha_j}$. These products form orthonormal Gaussian
$L_2$ bases, and rotations preserve total degree and norm. Since $\nu$
contains only degrees at least two, $T_a(Mz)=MB_az$ is orthogonal to
$T_a\nu$ and
$\norm{T_a\nu}_{L_2}^2\leq\|B_a\|_{\mathrm{op}}^4W$.
Write $\Sigma_B=\E_A[B_AB_A^\top]$, $\beta_B=\E_A\|B_A\|_{\mathrm{op}}^4$,
and $\mathcal L^\star=n-\tr\Sigma_B$.
Conditioning on $(z,A)$ gives
\begin{align*}
 \mathcal L
 &\geq n-\E_A\norm{T_Ah}_{L_2}^2
 \geq n-\tr(M\Sigma_BM^\top)-\beta_BW\\
 &=\mathcal L^\star+\tr\{\Sigma_B(I_n-M^\top M)\}-\beta_BW
 \geq\mathcal L^\star+\kappa_BW.
\end{align*}
The feasible pair $(h_0,J_{\omega_0})$ has residual
$Q_0(z'-B_Az)$, whose expected squared norm is $\mathcal L^\star$.
Thus every optimum has $W=0$; whitening makes $M=Q$ orthogonal.
Substitution gives the stated Gaussian conditional law.

\emph{Transition estimation.} For each action with positive probability,
its sample count diverges almost surely. Let $S_{m,a},R_{m,a}$ average
$yy^\top,y'y'^\top$ within action $a$. The strong law gives
$\widehat C_{m,a}\to C_a:=QB_aQ^\top$ and $S_{m,a},R_{m,a}\to I_n$
almost surely. Expanding the fitted residual moment,
\[
 \widehat V_{m,a}=R_{m,a}-2\widehat C_{m,a}\widehat C_{m,a}^\top
 +\widehat C_{m,a}S_{m,a}\widehat C_{m,a}^\top
 \longrightarrow I_n-C_aC_a^\top.
\]
Gaussian parameter convergence gives weak convergence for each fixed $y$.
If $\|B_a\|_{\mathrm{op}}<1$, the limiting covariance is positive definite,
as are the estimated covariances eventually. Their densities converge
pointwise, so Scheff\'e's lemma \citep{scheffe1947} gives total-variation convergence.
\end{proofE}

Under the stated assumptions, every global minimizer of the whitened population squared-error objective recovers the physical state up to an orthogonal transformation. Holding this encoder fixed, empirical transition-moment estimates consistently recover the action-conditioned Gaussian successor law. This joint-action extension of \citet{Klindt2026} motivates predictive representations for centralized world modeling.

\section{Method}
\label{sec:method}
We introduce \method{}, a joint-embedding predictive model-based MARL
algorithm for centralized training with decentralized execution. Each agent
maintains a local latent world model and acts from its own inferred state.
During training, a joint predictor models the effect of synchronized agent
states and actions by predicting each agent's next observation embedding.
These predictions are passed through the same local posterior used for real
observations, enabling joint imagination without constructing a centralized
actor state. A centralized critic is used for policy learning.

\subsection{Local Predictive World Model}
\label{sec:local-model}
\begin{wrapfigure}{R}{0.26\textwidth}
\centering
\includegraphics[width=0.92\linewidth]{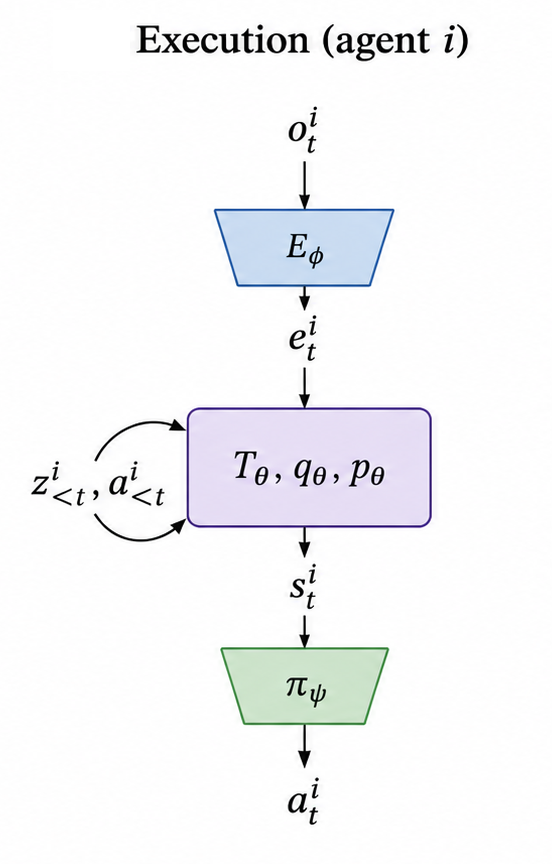}
\vspace{0.15em}
{\footnotesize
\raggedright
\textbf{Decentralized execution.}
Agent $i$ encodes its local observation and updates
$\mathbf s_t^i$ from its own observation--action history.
The shared policy then selects
$\mathbf a_t^i\sim\pi_\psi(\cdot\mid\mathbf s_t^i)$
without access to other agents.
\par}
\end{wrapfigure}

We denote the observation encoder by $E_\phi$ and the local latent dynamics
by parameters $\theta$, comprising a causal Transformer $T_\theta$, a
history-conditioned prior $p_\theta$, and an observation-conditioned posterior
$q_\theta$. Each agent's action-observation history is summarized by a deterministic history state $\mathbf h_t^i$ and a categorical stochastic latent state $\mathbf z_t^i$, with
\begin{equation}
\mathbf s_t^i
=
\operatorname{cat}\!\left(
\mathbf h_t^i,
\operatorname{vec}(\mathbf z_t^i)
\right)
\label{eq:local-feature}
\end{equation}
forming the local state used by the policy.

\paragraph{Local state inference.}
The encoder represents the current observation, while the causal Transformer summarizes the preceding stochastic states and actions. The posterior combines the observation embedding and history state to infer the current stochastic latent state.
\begin{equation}
\begin{aligned}
\mathbf e_t^i
&=E_\phi(\mathbf o_t^i),\\
\mathbf h_t^i
&=T_\theta(\mathbf z_{<t}^i,\mathbf a_{<t}^i),\\
\mathbf z_t^i
&\sim
q_\theta(
\cdot\mid\mathbf h_t^i,\mathbf e_t^i).
\end{aligned}
\label{eq:local-state}
\end{equation}
The prior $p_\theta(\mathbf z_t^i\mid\mathbf h_t^i)$ predicts the stochastic
state from local history alone and is trained jointly with the posterior.

The deterministic state $\mathbf h_{t}^i$ summarizes information from the preceding latent–action history, whereas $\mathbf z_t^i$
incorporates information revealed by the current observation. This separation allows the local dynamics to advance without observing the next environment input while retaining an observation-conditioned correction when one becomes available. Parameters are shared across agents, while latent histories are
maintained independently.

\paragraph{Learning joint-embeddings.}
We train two local predictors toward the EMA target of the embedding
$\bar{\mathbf e}_t^i$ defined in~\Cref{sec:jepa-background}.
The posterior predictor $P_{\mathrm{post}}$ maps the complete local state
$\mathbf s_t^i$ into the target embedding space, while the dynamics predictor
$P_{\mathrm{dyn}}$ maps only the deterministic history state
$\mathbf h_t^i$ into the same space:
\begin{equation}
\mathcal L_{\mathrm{post}}
=
d_{\cos}\!\left(
P_{\mathrm{post}}(\mathbf s_t^i),
\bar{\mathbf e}_t^i
\right),
\qquad
\mathcal L_{\mathrm{dyn}}
=
d_{\cos}\!\left(
P_{\mathrm{dyn}}(\mathbf h_t^i),
\bar{\mathbf e}_t^i
\right).
\label{eq:local-jepa}
\end{equation}

The two predictors impose complementary constraints on the local
representation. $P_{\mathrm{post}}$ is evaluated after the current observation
has been incorporated and therefore encourages $\mathbf s_t^i$ to retain the
information represented by the target encoder. In contrast,
$P_{\mathrm{dyn}}$ receives only $\mathbf h_t^i$ and must predict the same
target from the preceding latent--action history. The resulting representation
is therefore both descriptive of the current observation and predictable from
local dynamics. Crucially, this embedding space later forms the interface
through which the joint model supplies predicted observations to the local
world model.

The complete local objective additionally contains prior--posterior latent regularization, local action-availability prediction, and SIGReg. Reward and continuation are predicted by the joint model. The appendix gives the full objective for the vector-observation SMAC configuration.

\paragraph{Decentralized policy.}
Each agent acts only from its local latent state,
\begin{equation}
\mathbf a_t^i
\sim
\pi_\psi(\cdot\mid\mathbf s_t^i).
\label{eq:local-policy}
\end{equation}
Policy parameters are shared across agents, and each agent maintains its own local history.
\subsection{Joint Prediction of Multi-Agent Dynamics}
\label{sec:joint-model}

\begin{wrapfigure}{R}{0.29\textwidth}
\centering

\includegraphics[width=0.94\linewidth]
{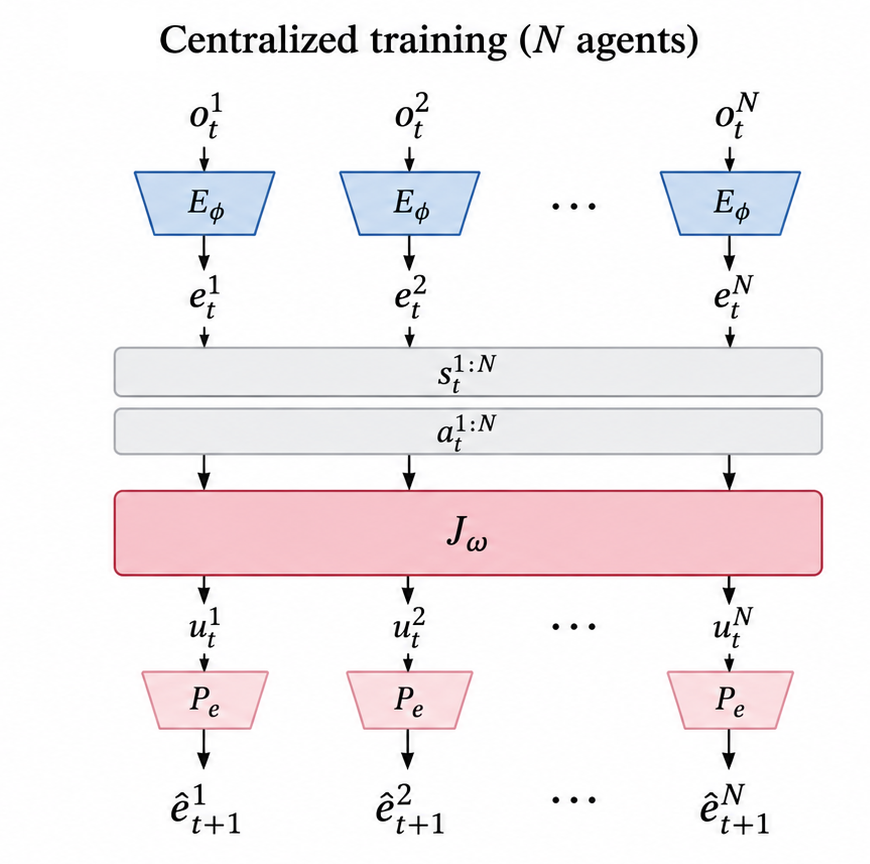}

\vspace{0.15em}
{\footnotesize
\raggedright
\textbf{Centralized training.}
The joint model $J_\omega$ combines synchronized local states
$\mathbf s_t^{1:N}$ with the joint action $\mathbf a_t^{1:N}$,
forming agent-specific interaction representations $\mathbf u_t^i$
from which it predicts the next local embeddings
$\widehat{\mathbf e}_{t+1}^i$.
\par}
\end{wrapfigure}

Local transition prediction does not account for the simultaneous actions of
other agents. During centralized training, we therefore introduce a joint
predictor $J_\omega$ conditioned on the synchronized local states
$\mathbf s_t^{1:N}$ and joint action $\mathbf a_t^{1:N}$. It produces an
agent-specific interaction representation,
\begin{equation}
\mathbf u_t^{1:N}
=
J_\omega\!\left(
\mathbf s_t^{1:N},
\mathbf a_t^{1:N}
\right),
\label{eq:joint-state}
\end{equation}
from which
\begin{equation}
\widehat{\mathbf e}_{t+1}^i
=
P_e(\mathbf u_t^i)
\label{eq:joint-prediction}
\end{equation}
predicts the next local observation embedding of agent $i$.

Conditioning on the joint action allows the predictor to model interaction
effects that cannot be inferred from the focal agent's local history alone.
The output remains agent-specific: the model predicts the representation of
each agent's next local observation rather than constructing a shared actor
state.
\paragraph{Joint predictor training.}
The predicted embedding is supervised by both the EMA target and the stopped
online encoding
\begin{equation}
\begin{aligned}
\mathcal L_{\mathrm{emb}}
&=
d_{\cos}\!\left(
\widehat{\mathbf e}_{t+1}^i,
\bar{\mathbf e}_{t+1}^i
\right),\\
\mathcal L_{\mathrm{int}}
&=
\operatorname{SmoothL1}\!\left(
\widehat{\mathbf e}_{t+1}^i,
\operatorname{sg}(E_\phi(\mathbf o_{t+1}^i))
\right).
\end{aligned}
\label{eq:joint-embedding-loss}
\end{equation}
The two targets serve different roles. The EMA target aligns joint prediction
with the representation learned by the local JEPA objectives. The stopped
online encoding matches the real encoding that the local posterior receives
during environment interaction. This objective reduces the mismatch between
encoded observations and joint-model predictions before both are consumed by
the same posterior during imagination.

Together with auxiliary transition predictions, the joint objective is
\begin{equation}
\mathcal L_{\mathrm{joint}}
=
\beta_e\mathcal L_{\mathrm{emb}}
+
\beta_I\mathcal L_{\mathrm{int}}
+
\mathcal L_{\mathrm{aux}}^J .
\label{eq:joint-objective}
\end{equation}
Here, $\mathcal L_{\mathrm{aux}}^J$ includes reward, continuation, agent-alive, and legal-action prediction. In the full method, factual reward, continuation, and agent-alive losses also update the local input representations at scale 1.0. The factual one-step embedding loss supplies an additional local-input gradient at scale 0.1. All other joint losses stop at those inputs. The joint predictor and its heads retain their ordinary loss weights. The baseline sets both additional gradient scales to zero.

\paragraph{Multi-step prediction.}
We supervise future embeddings directly from replay of realized past states. Let $\mathcal H$ denote the set of prediction horizons and
$\boldsymbol\chi_{t,h}^i$ the intervening action context used to predict the
future of agent $i$. For each $h\in\mathcal H$, a head predicts
\begin{equation}
\widehat{\mathbf e}_{t+h\mid t}^i
=
P_h\!\left(
\mathbf u_t^i,
\boldsymbol\chi_{t,h}^i
\right),
\qquad
\mathcal L_{\mathrm{MS}}
=
\sum_{h\in\mathcal H}
\alpha_h\,
d_{\cos}\!\left(
\widehat{\mathbf e}_{t+h\mid t}^i,
\bar{\mathbf e}_{t+h}^i
\right).
\label{eq:multistep-loss}
\end{equation}
We train each multi-step head to predict future embeddings from the recorded focal-agent action sequence. For horizons greater than one, an action-discrimination loss substitutes each distinct legal alternative for the final focal-agent action while keeping the joint root and earlier actions fixed. It encourages the factual prediction to exceed each alternative's cosine similarity to the observed future embedding by a margin of 0.1. The loss has coefficient 0.1 and favors distinct predictions even for actions with equivalent outcomes. The appendix gives the construction of $\boldsymbol\chi_{t,h}^i$ and the objective. These heads provide auxiliary training supervision. Imagination uses the recurrent one-step joint predictor.

\subsection{Joint Imagination and Decentralized Policy Learning}
\label{sec:joint-imagination}

\paragraph{Shared posterior interface.}
During imagination, agents sample actions from their local policies and advance
their local deterministic states. The joint predictor then produces the next
observation embedding from the synchronized imagined states and joint action
\begin{equation}
\begin{aligned}
\widehat{\mathbf a}_t^i
&\sim
\pi_\psi(\cdot\mid\widehat{\mathbf s}_t^i),\\
\widehat{\mathbf h}_{t+1}^i
&=
T_\theta(
\widehat{\mathbf z}_{\leq t}^i,
\widehat{\mathbf a}_{\leq t}^i),\\
\widehat{\mathbf e}_{t+1}^i
&=
P_e\!\left(
J_\omega(
\operatorname{sg}(\widehat{\mathbf s}_t^{1:N}),
\widehat{\mathbf a}_t^{1:N})_i
\right).
\end{aligned}
\label{eq:imagined-transition}
\end{equation}
The predicted embedding replaces the encoder output in the same posterior used
for real observations
\begin{equation}
\begin{aligned}
\mathbf z_{t+1}^i
&\sim
q_\theta(
\cdot\mid
\mathbf h_{t+1}^i,
E_\phi(\mathbf o_{t+1}^i)),
\\
\widehat{\mathbf z}_{t+1}^i
&\sim
q_\theta(
\cdot\mid
\widehat{\mathbf h}_{t+1}^i,
\widehat{\mathbf e}_{t+1}^i).
\end{aligned}
\label{eq:shared-posterior}
\end{equation}
The joint predictor therefore does not generate the next actor state directly.
It predicts only the observation-side input to the local posterior. The
posterior combines this prediction with the independently advanced local
dynamics state in the same manner as it combines a real encoded observation
with $\mathbf h_{t+1}^i$. Consequently, centralized information changes the
predicted consequence of the joint action without changing the mechanism that
constructs the decentralized policy state.

\paragraph{Centralized value learning.}
The critic uses a central value network $V_{\xi}$ that takes as inputs the stopped local states of all agents during training,
\begin{equation}
V_\xi^i
=
V_\xi\!\left(
\operatorname{sg}(\widehat{\mathbf s}_t^{1:N})
\right)_i.
\label{eq:central-value}
\end{equation}
Using rewards and discounted continuations predicted by the joint model, we compute imagined $\lambda$-returns from \eqref{eq:prelim-lambda-return}. The centralized critic attends to the agents' local latent states together with roster and liveness masks. Each actor conditions on its own $\mathbf s_t^i$. The actor is optimized from the imagined returns and the critic from the corresponding value targets, with gradients stopped at the world-model states.

At execution time, each agent uses only the shared local encoder, latent
dynamics, posterior, and policy. The joint predictor, direct multi-step
objectives, and centralized critic are restricted to training.

\section{Experiments}
\label{sec:experiments}

\subsection{Experimental Setup}
We evaluate whether joint-embedding world models support effective policy learning from latent imagination in the StarCraft Multi-Agent Challenge (SMAC, \citealp{samvelyan2019starcraft}). The main comparison covers eight SMAC v1 tasks under StarCraft II 4.10.0, including homogeneous and heterogeneous teams, asymmetric engagements, and different coordination demands. We further perform various gradient ablations. A detailed description of SMAC can be found in Appendix~\ref{app:smac}.

We compare MA-JEPA against DMAWM, MAPPO, QMIX, and MAT at matched interaction budgets. MA-JEPA hyperparameters are given in
Appendix~\ref{app:majepa-hyperparameters}, and baseline configurations in
Appendix~\ref{app:baseline-configuration}. MAPPO is additionally trained to
1M steps as a higher-budget reference. All algorithms were benchmarked under NVIDIA A100 PCIe GPUs.

\subsection{Results}
\label{sec:smac-results}

\begin{figure*}[!htbp]
\centering
\includegraphics[width=\textwidth]{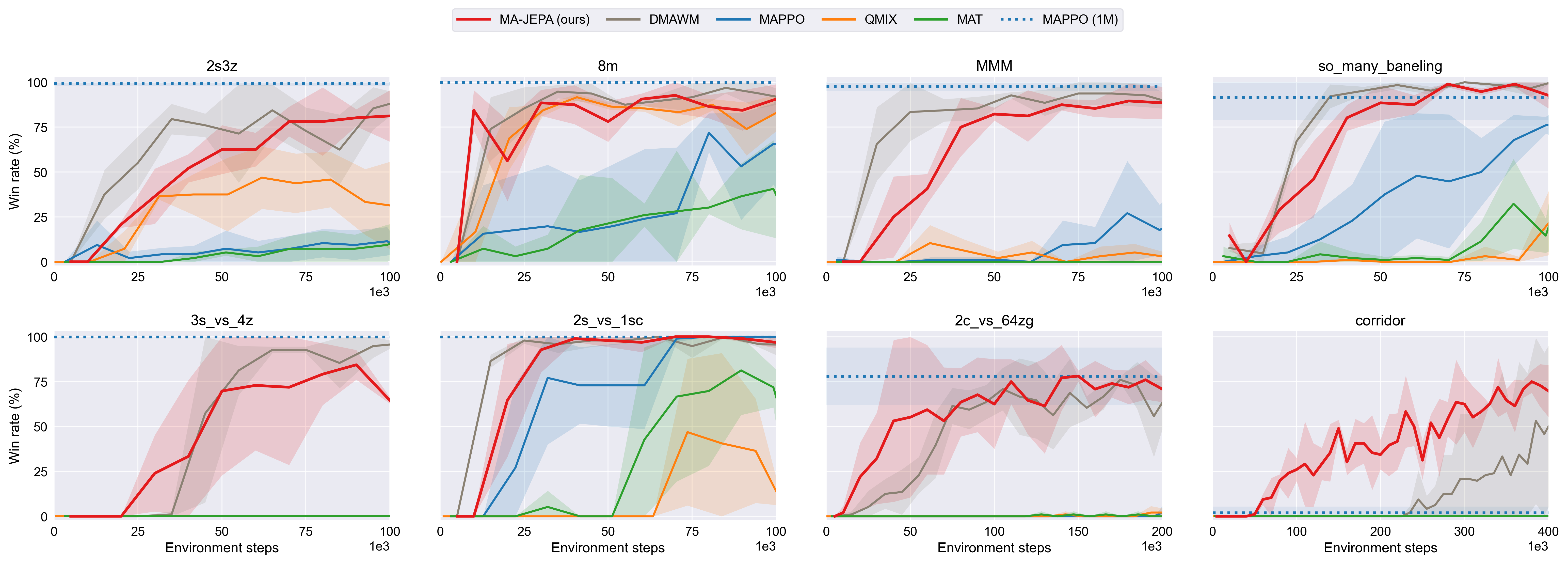}
\caption{
Win rate versus environment steps on eight SMAC v1 tasks at the indicated
budgets. Solid curves show means across available training seeds, with
$\pm1$ sample-standard-deviation bands.
Recorded evaluations are connected by linear interpolation without smoothing; The dotted horizontal line shows MAPPO's mean win rate at 1M steps.}
\label{fig:smac-learning-curves}
\end{figure*}

\Cref{tab:smac_matched_budget} reports win rates, and
\Cref{fig:smac-learning-curves} shows learning curves for the eight tasks.
These results demonstrate effective control through joint-embedding-based imagination.

Performance remains task-dependent. We match or achieve the highest reported mean win rate on four of the eight tasks. On \texttt{so\_many\_baneling} and \texttt{2s\_vs\_1sc}, win rates approach 100\%, closely matching those of DMAWM (within 1 SD of those results). Our results on \texttt{3s\_vs\_4z} and the similar \texttt{3s\_vs\_5z} were noticeably weaker and had significantly higher variance. Agents were able to learn near optimal solutions on some of the seeds, but this was offset by poor performance on others. This is also true, though, to a lesser extent, for DMAWM. Our current hypothesis is that this high-variation in training results because these settings specifically study ``kiting'' behavior, where to win an agent must repeatedly approach, hit, and run away from the enemy. This is a multi-step sequential task that is highly recurrent throughout the episode and may be many steps beyond the rollout length of our current world model during imagination training. We therefore further hypothesize this sequence of actions is not currently captured effectively in the latent space of the world model consistently, leading the policy to fail to train.

In follow-up work, we plan to explore the geometry of the latent space embedding to understand exactly what is and is not learned by the world model in these ``kiting'' settings, what the optimal solution looks like, and identify optimal and learned policy features like recurrent action sequences.

\definecolor{smacEasy}{HTML}{246A6B}
\definecolor{smacHard}{HTML}{A65324}
\definecolor{smacSuperHard}{HTML}{78466D}
\providecommand{\smacEasyColor}[1]{\textcolor{smacEasy}{\textbf{#1}}}
\providecommand{\smacHardColor}[1]{\textcolor{smacHard}{\textbf{#1}}}
\providecommand{\smacSuperHardColor}[1]{\textcolor{smacSuperHard}{\textbf{#1}}}
\providecommand{\res}[2]{#1\,(#2)}
\providecommand{\best}[2]{\textbf{#1}\,(#2)}
\newcommand{\smacgroup}[3]{%
  \makebox[0pt][r]{%
    \multirow[c]{#1}{*}{%
      \raisebox{\dimexpr(\ht\strutbox-\dp\strutbox+\depth-\height)/2\relax}[0pt][0pt]{%
        \rotatebox[origin=c]{90}{%
          \textcolor{#2}{\scriptsize\bfseries #3}}}}%
    \hspace{\dimexpr\tabcolsep+0.4em\relax}%
  }%
}

\begin{table*}[t]
\centering
\caption{SMAC win rate (\%) at matched nominal environment-step budgets,
reported as mean (sample standard deviation) across available training
seeds from the set 1302, 2771, and 7636. Results use 100-episode
evaluations at the budget checkpoint. \textbf{Bold} denotes the highest
available mean win rate for each map. Rows are grouped into \smacEasyColor{Easy}, \smacHardColor{Hard},
and \smacSuperHardColor{Super Hard} maps.}
\label{tab:smac_matched_budget}
\resizebox{\textwidth}{!}{%
\begin{tabular}{lrccccc}
\toprule
Map & Steps & \method{} (ours) & DMAWM & MAPPO & QMIX & MAT \\
\midrule
\smacgroup{5}{smacEasy}{Easy}%
\texttt{2s\_vs\_1sc} & 100k
& \res{97.3}{2.5} & \res{95.7}{5.9} & \best{100.0}{0.0}
& \res{13.7}{7.4} & \res{51.3}{27.1} \\
\texttt{2s3z} & 100k
& \res{78.7}{14.5} & \best{88.0}{7.0} & \res{8.0}{2.6}
& \res{31.3}{24.5} & \res{16.5}{9.2} \\
\texttt{8m} & 100k
& \best{92.0}{1.0} & \best{92.0}{4.6} & \res{65.7}{33.1}
& \res{83.0}{10.1} & \res{26.0}{22.6} \\
\texttt{MMM} & 100k
& \best{91.0}{3.0} & \res{90.0}{4.6} & \res{20.7}{17.0}
& \res{3.0}{2.6} & \res{0.0}{0.0} \\
\texttt{so\_many\_baneling} & 100k
& \res{95.0}{6.1} & \best{99.5}{0.7} & \res{77.0}{7.2}
& \res{21.7}{18.0} & \res{32.0}{24.3} \\
\midrule
\smacgroup{2}{smacHard}{Hard}%
\texttt{3s\_vs\_4z} & 100k
& \res{74.0}{7.5} & \best{95.7}{2.3} & \res{0.0}{0.0}
& \res{0.0}{0.0} & \res{0.0}{0.0} \\
\texttt{2c\_vs\_64zg} & 200k
& \best{78.3}{9.5} & \res{75.3}{8.9} & \res{0.0}{0.0}
& \res{0.0}{0.0} & \res{0.0}{0.0} \\
\midrule
\smacgroup{1}{smacSuperHard}{\shortstack{Super\\[-2pt]Hard}}%
\texttt{corridor} & 400k
& \best{66.0}{3.5} & \res{50.0}{44.9} & \res{0.0}{0.0}
& \res{0.0}{0.0} & \res{0.0}{0.0} \\
\bottomrule
\end{tabular}%
}
\end{table*}

\subsection{Ablations}
\label{sec:ablations}

Figure~\ref{app:gradient-ablations} shows that the effects of the two
gradient routes depend on the task. Removing factual JEPA gradients
reduces performance on \texttt{2s3z} but improves it on
\texttt{3s\_vs\_4z}. Conversely, removing outcome gradients improves
performance on \texttt{2s3z}, but yields lower mean performance and
substantially greater seed variability on \texttt{3s\_vs\_4z}.
Enabling both routes provides a more balanced configuration across
these tasks, although it does not achieve the highest mean on either
task individually.
We isolate how centralized prediction losses shape the representations used
by decentralized policies. All four variants retain the local world model,
joint predictor, outcome heads, and latent-imagination policy training.
The ablation changes which factual joint losses additionally backpropagate
into the local features supplied to the joint model.

We vary two gradient routes in a $2\times2$ factorial design.
The \emph{outcome route} propagates gradients from factual reward,continuation, and agent-alive prediction into the local input features.
The \emph{JEPA route} propagates the factual joint embedding-prediction
error into those features at scale 0.1. This error compares the predicted
embedding with a stopped exponential-moving-average target using cosine
loss. The scale applies to the additional local-input gradient, not to the
ordinary training loss of the joint predictor.

All four variants retain the action-discrimination loss with margin 0.1 and loss coefficient 0.1.

\section{Conclusion and Future Work}
\label{sec:conclusion}

We introduced \method{}, a joint-embedding predictive world model for
multi-agent reinforcement learning under centralized training and decentralized
execution. Rather than reconstructing observations, the local world model
learns predictive representations that support latent imagination and
decentralized control. A training-only joint model captures interactions
between agents by predicting each agent's next local observation embedding,
which is processed by the same local posterior used during real interaction.

Our experiments show that this architecture performs strongly on SMAC at budgets of 100k, 200k, and 400k environment steps, matching or exceeding the strongest reported comparator mean win rate on four of eight evaluated maps.

These results suggest that joint-embedding prediction provides a viable alternative to reconstructive world models for multi-agent control. An important direction for future work is to understand which properties of the learned representations make them effective for world-model prediction and control. In particular, we are interested in whether predictive objectives recover representations that separate persistent physical dynamics, agent identity, and interaction-dependent information, and how such structure affects generalization across environments and teams. Extending \method{} to broader continuous-control and heterogeneous multi-agent settings is another natural direction.

\bibliographystyle{plainnat}
\bibliography{references}

@article{scheffe1947,
  author  = {Scheff{\'e}, Henry},
  title   = {A Useful Convergence Theorem for Probability Distributions},
  journal = {The Annals of Mathematical Statistics},
  year    = {1947},
  volume  = {18},
  number  = {3},
  pages   = {434--438},
  doi     = {10.1214/aoms/1177730390},
  url     = {https://doi.org/10.1214/aoms/1177730390}
}

@misc{Klindt2026,
  title         = {When Does {LeJEPA} Learn a World Model?},
  author        = {Klindt, David and LeCun, Yann and Balestriero, Randall},
  year          = {2026},
  eprint        = {2605.26379},
  archivePrefix = {arXiv},
  primaryClass  = {stat.ML},
  doi={10.48550/arXiv.2605.26379},
  url= {https://arxiv.org/abs/2605.26379}
}

@inproceedings{Xue2026,
  title     = {Learning Disentangled Multi-Agent World Model for Decentralized Control},
  author    = {Xue, Di and Jiang, Jing and Zhang, Shaowei and Guo, Wenhao and
               Yuan, Lei and Zhang, Zongzhang and Yu, Yang},
  booktitle = {Forty-Third International Conference on Machine Learning},
  year      = {2026},
  url       = {https://openreview.net/forum?id=nYyfpPubnW}
}

@article{kraemer2016,
title = {Multi-agent reinforcement learning as a rehearsal for decentralized planning},
journal = {Neurocomputing},
volume = {190},
pages = {82-94},
year = {2016},
issn = {0925-2312},
doi = {https://doi.org/10.1016/j.neucom.2016.01.031},
url = {https://www.sciencedirect.com/science/article/pii/S0925231216000783},
author = {Landon Kraemer and Bikramjit Banerjee}
}

@article{oliehoek2008,
author = {Oliehoek, Frans A. and Spaan, Matthijs T. J. and Vlassis, Nikos},
title = {Optimal and approximate Q-value functions for decentralized POMDPs},
year = {2008},
issue_date = {May 2008},
publisher = {AI Access Foundation},
address = {El Segundo, CA, USA},
volume = {32},
number = {1},
issn = {1076-9757},
journal = {J. Artif. Int. Res.},
month = may,
pages = {289–353},
numpages = {65}
}

@article{sutton1988td,
  title   = {Learning to Predict by the Methods of Temporal Differences},
  author  = {Sutton, Richard S.},
  journal = {Machine Learning},
  volume  = {3},
  pages   = {9--44},
  year    = {1988}
}

@misc{lecun2022path,
  title  = {A Path Towards Autonomous Machine Intelligence},
  author = {LeCun, Yann},
  year   = {2022},
  note   = {Version 0.9.2},
  url    = {https://openreview.net/forum?id=BZ5a1r-kVsf}
}

@inproceedings{assran2023ijepa,
  title     = {Self-Supervised Learning from Images with a Joint-Embedding Predictive Architecture},
  author    = {Assran, Mahmoud and Duval, Quentin and Misra, Ishan and Bojanowski, Piotr and Vincent, Pascal and Rabbat, Michael and LeCun, Yann and Ballas, Nicolas},
  booktitle = {Proceedings of the IEEE/CVF Conference on Computer Vision and Pattern Recognition},
  pages     = {15619--15629},
  year      = {2023}
}

@inproceedings{grill2020byol,
  title     = {Bootstrap Your Own Latent: A New Approach to Self-Supervised Learning},
  author    = {Grill, Jean-Bastien and Strub, Florian and Altch{\'e}, Florent and Tallec, Corentin and Richemond, Pierre H. and Buchatskaya, Elena and Doersch, Carl and Avila Pires, Bernardo and Guo, Zhaohan Daniel and Azar, Mohammad Gheshlaghi and Piot, Bilal and Kavukcuoglu, Koray and Munos, R{\'e}mi and Valko, Michal},
  booktitle = {Advances in Neural Information Processing Systems},
  volume    = {33},
  pages     = {21271--21284},
  year      = {2020}
}

@article{bardes2024vjepa,
  title   = {Revisiting Feature Prediction for Learning Visual Representations from Video},
  author  = {Bardes, Adrien and Garrido, Quentin and Ponce, Jean and Chen, Xinlei and Rabbat, Michael and LeCun, Yann and Assran, Mido and Ballas, Nicolas},
  journal = {Transactions on Machine Learning Research},
  year    = {2024}
}

@misc{balestriero2025lejepa,
  title         = {{LeJEPA}: Provable and Scalable Self-Supervised Learning Without the Heuristics},
  author        = {Balestriero, Randall and LeCun, Yann},
  year          = {2025},
  eprint        = {2511.08544},
  archivePrefix = {arXiv},
  primaryClass  = {cs.LG}
}

@misc{assran2025vjepa2,
  title  = {{V-JEPA 2}: Self-Supervised Video Models Enable Understanding, Prediction and Planning},
  author = {
    Assran, Mahmoud and
    Bardes, Adrien and
    Fan, David and
    Garrido, Quentin and
    Howes, Russell and
    Komeili, Mojtaba and
    Muckley, Matthew and
    Rizvi, Ammar and
    Roberts, Claire and
    Sinha, Koustuv and
    Zholus, Artem and
    Arnaud, Sergio and
    Gejji, Abha and
    Martin, Ada and
    Robert Hogan, Francois and
    Dugas, Daniel and
    Bojanowski, Piotr and
    Khalidov, Vasil and
    Labatut, Patrick and
    Massa, Francisco and
    Szafraniec, Marc and
    Krishnakumar, Kapil and
    Li, Yong and
    Ma, Xiaodong and
    Chandar, Sarath and
    Meier, Franziska and
    LeCun, Yann and
    Rabbat, Michael and
    Ballas, Nicolas
  },
  year          = {2025},
  eprint        = {2506.09985},
  archivePrefix = {arXiv},
  primaryClass  = {cs.AI}
}

@misc{maes2026leworldmodel,
  title         = {{LeWorldModel}: Stable End-to-End Joint-Embedding Predictive Architecture from Pixels},
  author        = {Maes, Lucas and Le Lidec, Quentin and Scieur, Damien and LeCun, Yann and Balestriero, Randall},
  year          = {2026},
  eprint        = {2603.19312},
  archivePrefix = {arXiv},
  primaryClass  = {cs.LG}
}

@misc{bagatella2025tdjepa,
  title         = {{TD-JEPA}: Latent-Predictive Representations for Zero-Shot Reinforcement Learning},
  author        = {Bagatella, Marco and Pirotta, Matteo and Touati, Ahmed and Lazaric, Alessandro and Tirinzoni, Andrea},
  year          = {2025},
  eprint        = {2510.00739},
  archivePrefix = {arXiv},
  primaryClass  = {cs.LG}
}

@article{moerland2023mbrl,
  title   = {Model-Based Reinforcement Learning: A Survey},
  author  = {Moerland, Thomas M. and Broekens, Joost and Plaat, Aske and Jonker, Catholijn M.},
  journal = {Foundations and Trends in Machine Learning},
  volume  = {16},
  number  = {1},
  pages   = {1--118},
  year    = {2023},
  doi     = {10.1561/2200000086}
}

@inproceedings{hafner2019planet,
  title     = {Learning Latent Dynamics for Planning from Pixels},
  author    = {Hafner, Danijar and Lillicrap, Timothy and Fischer, Ian and Villegas, Ruben and Ha, David and Lee, Honglak and Davidson, James},
  booktitle = {Proceedings of the 36th International Conference on Machine Learning},
  series    = {Proceedings of Machine Learning Research},
  volume    = {97},
  pages     = {2555--2565},
  publisher = {PMLR},
  year      = {2019}
}

@inproceedings{hafner2019dream,
  title     = {Dream to Control: Learning Behaviors by Latent Imagination},
  author    = {Hafner, Danijar and Lillicrap, Timothy and Ba, Jimmy and Norouzi, Mohammad},
  booktitle = {International Conference on Learning Representations},
  year      = {2020}
}

@article{hafner2023dreamerv3,
  title     = {Mastering Diverse Control Tasks Through World Models},
  author    = {Hafner, Danijar and Pasukonis, Jurgis and Ba, Jimmy and Lillicrap, Timothy},
  journal   = {Nature},
  volume    = {640},
  number    = {8059},
  pages     = {647--653},
  year      = {2025},
  doi       = {10.1038/s41586-025-08744-2}
}

@inproceedings{micheli2023iris,
  title     = {Transformers are Sample-Efficient World Models},
  author    = {Micheli, Vincent and Alonso, Eloi and Fleuret, Fran{\c{c}}ois},
  booktitle = {International Conference on Learning Representations},
  year      = {2023}
}

@inproceedings{zhang2023storm,
  title     = {{STORM}: Efficient Stochastic Transformer Based World Models for Reinforcement Learning},
  author    = {Zhang, Weipu and Wang, Gang and Sun, Jian and Yuan, Yetian and Huang, Gao},
  booktitle = {Advances in Neural Information Processing Systems},
  volume    = {36},
  year      = {2023}
}

@inproceedings{janner2019mbpo,
  title     = {When to Trust Your Model: Model-Based Policy Optimization},
  author    = {Janner, Michael and Fu, Justin and Zhang, Marvin and Levine, Sergey},
  booktitle = {Advances in Neural Information Processing Systems},
  volume    = {32},
  year      = {2019}
}

@inproceedings{egorov2022mamba,
  title     = {Scalable Multi-Agent Model-Based Reinforcement Learning},
  author    = {Egorov, Vladimir and Shpilman, Aleksei},
  booktitle = {Proceedings of the 21st International Conference on Autonomous Agents and Multiagent Systems},
  pages     = {381--390},
  year      = {2022}
}

@inproceedings{venugopal2024mabl,
  title     = {{MABL}: Bi-Level Latent-Variable World Model for Sample-Efficient Multi-Agent Reinforcement Learning},
  author    = {Venugopal, Aravind and Milani, Stephanie and Fang, Fei and Ravindran, Balaraman},
  booktitle = {Proceedings of the 23rd International Conference on Autonomous Agents and Multiagent Systems},
  pages     = {1865--1873},
  year      = {2024}
}

@article{zhang2024marie,
  title   = {Decentralized Transformers with Centralized Aggregation are Sample-Efficient Multi-Agent World Models},
  author  = {Zhang, Yang and Bai, Chenjia and Zhao, Bin and Yan, Junchi and Li, Xiu and Li, Xuelong},
  journal = {Transactions on Machine Learning Research},
  year    = {2025}
}

@misc{deihim2025matwm,
  title         = {Transformer World Model for Sample Efficient Multi-Agent Reinforcement Learning},
  author        = {Deihim, Azad and Alonso, Eduardo and Apostolopoulou, Dimitra},
  year          = {2025},
  eprint        = {2506.18537},
  archivePrefix = {arXiv},
  primaryClass  = {cs.LG}
}

@inproceedings{samvelyan2019starcraft,
  title     = {The {StarCraft} Multi-Agent Challenge},
  author    = {Samvelyan, Mikayel and Rashid, Tabish and Schroeder de Witt, Christian and Farquhar, Gregory and Nardelli, Nantas and Rudner, Tim G. J. and Hung, Chia-Man and Torr, Philip H. S. and Foerster, Jakob and Whiteson, Shimon},
  booktitle = {Proceedings of the 18th International Conference on Autonomous Agents and MultiAgent Systems},
  pages     = {2186--2188},
  year      = {2019}
}

@inproceedings{yu2022mappo,
  title     = {The Surprising Effectiveness of {PPO} in Cooperative Multi-Agent Games},
  author    = {Yu, Chao and Velu, Akash and Vinitsky, Eugene and
               Gao, Jiaxuan and Wang, Yu and Bayen, Alexandre and Wu, Yi},
  booktitle = {Advances in Neural Information Processing Systems},
  volume    = {35},
  pages     = {24611--24624},
  year      = {2022},
  doi       = {10.52202/068431-1787},
  url       = {https://proceedings.neurips.cc/paper_files/paper/2022/hash/9c1535a02f0ce079433344e14d910597-Abstract.html}
}

@inproceedings{wen2022mat,
  title     = {Multi-Agent Reinforcement Learning is a Sequence Modeling Problem},
  author    = {Wen, Muning and Kuba, Jakub Grudzien and Lin, Runji and
               Zhang, Weinan and Wen, Ying and Wang, Jun and Yang, Yaodong},
  booktitle = {Advances in Neural Information Processing Systems},
  volume    = {35},
  pages     = {16509--16521},
  year      = {2022},
  doi       = {10.52202/068431-1201},
  url       = {https://proceedings.neurips.cc/paper_files/paper/2022/hash/69413f87e5a34897cd010ca698097d0a-Abstract-Conference.html}
}

@inproceedings{rashid2018qmix,
  title     = {{QMIX}: Monotonic Value Function Factorisation for Deep Multi-Agent Reinforcement Learning},
  author    = {Rashid, Tabish and Samvelyan, Mikayel and
               Schroeder de Witt, Christian and Farquhar, Gregory and
               Foerster, Jakob and Whiteson, Shimon},
  booktitle = {Proceedings of the 35th International Conference on Machine Learning},
  series    = {Proceedings of Machine Learning Research},
  volume    = {80},
  pages     = {4295--4304},
  publisher = {PMLR},
  year      = {2018},
  url       = {https://proceedings.mlr.press/v80/rashid18a.html}
}

\appendix

\newpage
\section{Comparing Multi-Agent World Model Architectures}

\begin{table*}[!htpb]
    \centering
    \caption{Comparison of MA-JEPA with multi-agent world model architectures.}
    \label{tab:world_model_comparison}
    \renewcommand{\arraystretch}{1.10}
    \setlength{\tabcolsep}{5.5pt}
    \resizebox{\textwidth}{!}{%
    \begin{tabular}{lcccccc}
        \toprule
        \textbf{Aspect}
        & \textbf{MARIE}
        & \textbf{MAMBA}
        & \textbf{MBVD}
        & \textbf{MATWM}
        & \textbf{DMAWM}
        & \textbf{MA-JEPA} \\
        \midrule

        \textbf{Multi-Agent Design}
        & Yes
        & Yes
        & Yes
        & Yes
        & Yes
        & Yes \\

        \textbf{Backbone Architecture}
        & Transformer
        & GRU
        & GRU
        & Transformer
        & GRU + Transformer
        & Transformer \\

        \textbf{Latent Representation}
        & VQ-VAE
        & Categorical VAE
        & VAE
        & Categorical VAE
        & Categorical RSSM
        & \makecell{Categorical latent} \\

        \textbf{Critic Type}
        & Centralized
        & Centralized
        & Centralized
        & Semi-centralized
        & Centralized
        & Centralized \\

        \textbf{Agent Training}
        & PPO-style
        & PPO-style
        & Deep Q-learning
        & DreamerV3-style
        & PPO-style
        & \makecell{PPO-style} \\
        \bottomrule
    \end{tabular}%
    }
\end{table*}

\section{Environment and Task Details}
\label{app:environments}

We evaluate MA-JEPA on Version 1 of the Starcraft Multi-agent Challenge (SMAC)~\citep{samvelyan2019starcraft}.
Our task selection follows the evaluation settings used by closely related multi-agent world-model methods, enabling direct comparison under existing benchmark configurations.

\subsection{SMAC}
\label{app:smac}
SMAC consists of cooperative micromanagement scenarios in StarCraft II. Each allied unit is controlled by an individual agent and receives only a local observation of the battlefield. Observations contain information about visible allied and enemy units, including their relative positions, distances, health, shield values, and unit types. Agents select from discrete movement, stop, and unit-targeted attack actions. Invalid actions are excluded using the environment-provided legal-action mask.

SMAC is particularly suitable for evaluating multi-agent world models because an agent's local transition can depend strongly on the simultaneous actions of its teammates. For example, whether an enemy unit remains alive after an attack cannot in general be determined from the focal agent's action alone. MA-JEPA therefore uses the joint action during centralized world-model training while retaining a decentralized
policy conditioned only on each agent's local interaction history.

The main comparison covers eight SMAC scenarios, namely \texttt{2s\_vs\_1sc}, \texttt{2s3z},
\texttt{8m}, 
\texttt{MMM}, \texttt{so\_many\_baneling}, \texttt{3s\_vs\_4z}, 
\texttt{2c\_vs\_64zg},
and \texttt{corridor}. These scenarios cover homogeneous and heterogeneous teams and range from relatively simple coordination problems to difficult scenarios requiring precise joint behavior.

\begin{table}[!htbp]
    \centering
    \caption{SMAC scenarios used in our experiments.}
    \label{tab:smac_tasks}
    \small
    \begin{tabular}{lcl}
        \toprule
        \textbf{Scenario} & \textbf{Difficulty} & \textbf{Training budget} \\
        \midrule
        \texttt{2s\_vs\_1sc}        & \smacEasyColor{Easy}       & 100K \\
        \texttt{2s3z}              & \smacEasyColor{Easy}       & 100K \\
        \texttt{8m}                & \smacEasyColor{Easy} & 100K \\
        \texttt{MMM}               & \smacEasyColor{Easy}       & 100K \\
        \texttt{so\_many\_baneling} & \smacEasyColor{Easy}       & 100K \\
        \midrule
        \texttt{3s\_vs\_4z}         & \smacHardColor{Hard}       & 100K \\
        \texttt{2c\_vs\_64zg}       & \smacHardColor{Hard}       & 200K \\
        \midrule
        \texttt{corridor}          & \smacSuperHardColor{Super Hard} & 400K \\
        \bottomrule
    \end{tabular}
\end{table}

\newpage
\section{Additional Ablations}
\begin{figure*}[!htbp]
    \centering
    \includegraphics[width=\textwidth]{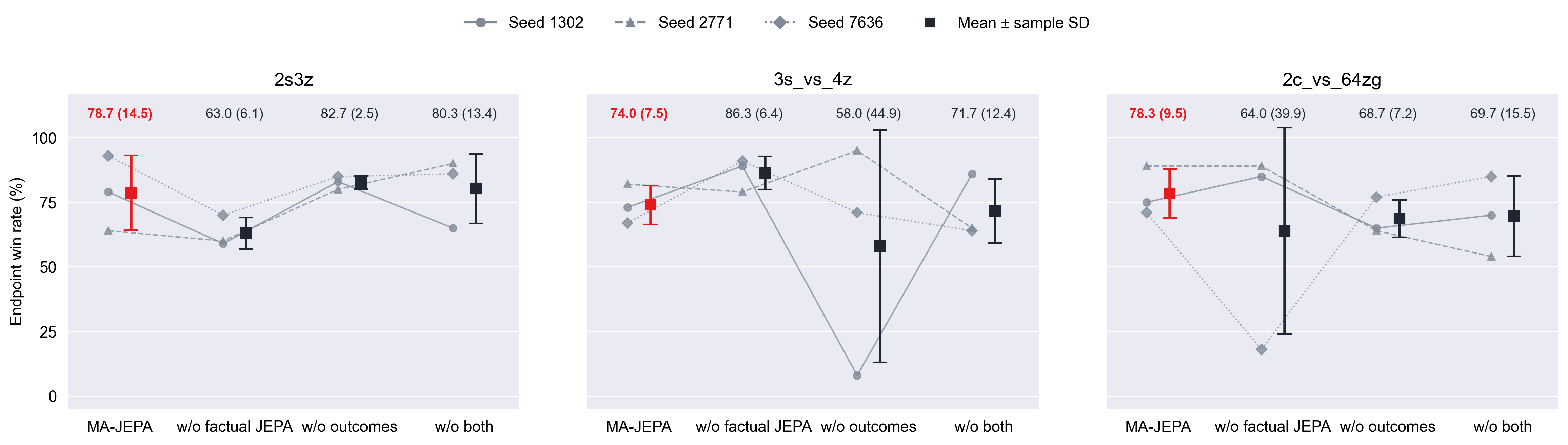}
    \caption{
    Ablation of gradient routes from joint prediction losses into local
    representations. MA-JEPA enables outcome gradients at scale 1.0
    and factual JEPA gradients at scale 0.1.
    ``w/o factual JEPA,'' ``w/o outcomes,'' and ``w/o both'' disable
    backpropogation from the local representations to the joint representation. In all cases, the joint representation is trained
    from our prediction objective.
    Connected markers show matched training seeds. Squares and error
    bars indicate the mean and sample standard deviation across three
    seeds, also reported numerically above each configuration.
    Results use 100-episode endpoint evaluations.
    }
    \label{app:gradient-ablations}
\end{figure*}

Across the three ablation maps in Figure~\ref{app:gradient-ablations},
MA-JEPA achieves the highest average win rate: 77.0\%, compared with
71.1\% without factual JEPA gradients, 69.8\% without outcome gradients,
and 73.9\% without either route. Although individual ablations perform
better on particular maps, their benefits do not transfer consistently
across tasks. Combining both routes therefore provides the strongest
overall performance across this evaluation set.

One possible explanation is that the two objectives provide complementary
training signals for local representations. Outcome gradients emphasize
features relevant to reward and termination, while factual JEPA gradients
encourage features that support prediction of future latent states.
Outcome supervision alone may leave useful information underconstrained, whereas predictive supervision alone need not prioritize
features relevant to control. Combining task-relevant outcome gradients with a weaker predictive gradient may balance these demands. The map-dependent results are consistent with this interpretation.

\section{Implementation Details}
\label{app:implementation}

\subsection{Replay Sampling}
\label{app:replay}

MA-JEPA maintains one replay buffer with two independently sampled views.
The world-model view supplies model supervision, while the behavior view supplies the roots for imagined policy learning. Both views use the same mixture of uniform and exponential-recency sampling
\begin{equation}
    P(v)
    =
    \eta \frac{1}{|\mathcal V|}
    +
    (1-\eta)
    \frac{d^{\operatorname{age}(v)}}
         {\sum_{w\in\mathcal V} d^{\operatorname{age}(w)}},
    \qquad
    \eta=0.5,\quad d=0.9998,
    \label{eq:replay-mixture}
\end{equation}
where $\mathcal V$ is the set of eligible sequence starts and age is
measured by insertion order. The two views use independent random
streams but may contain overlapping sequences. Independent sampling
therefore separates the two batches without removing recency bias from
the policy-root distribution.

Each replay sequence contains 192 context records followed by 64 learning records. The context records reconstruct the model’s temporal history; no loss is computed on them.

For world-model training, we reconstruct local temporal states from stored latent–action inputs and stochastic states using the current temporal network. For policy and value training, performed after the world-model update, we re-encode the context observations and infer their posterior states using the updated model. In both cases, the resulting local features and recorded joint actions reconstruct the joint Transformer’s temporal history.

Histories reset at episode boundaries. States supplied to policy and value learning are treated as fixed inputs, preventing these updates from changing the world model.

\paragraph{Sketched Isotropic Gaussian Regularization.}
To prevent representational collapse, we regularize the online observation
embeddings with SIGReg~\citep{balestriero2025lejepa}. For each agent slot,
SIGReg samples random unit directions $a_m$ and compares the empirical
one-dimensional projected embedding distribution
$u_{n,m}=a_m^\top e_n$ with a standard normal distribution using their
characteristic functions. For projection $m$, the discrepancy is

\begin{equation}
    \mathcal T_m
    =
    N
    \int
    \left|
        \frac{1}{N}\sum_{n=1}^{N}
        e^{\mathrm{i}t u_{n,m}}
        -
        e^{-t^2/2}
    \right|^2
    e^{-t^2/2}\,dt ,
\end{equation}

and $\mathcal L_{\mathrm{SIGReg}}$ is obtained by averaging this statistic
over random projections and nonempty agent slots. In practice, the integral
is approximated on a finite frequency grid. By encouraging many random
one-dimensional projections to resemble $\mathcal N(0,1)$, SIGReg promotes
approximately isotropic, non-collapsed embeddings without requiring an
observation reconstruction objective.

\subsection{Local World-Model Objective}
\label{app:local-objective}

Let $e_t^i=E_\phi(o_t^i)$ denote the online observation embedding and $\bar e_t^i=\operatorname{sg}(E_{\bar\phi}(o_t^i))$ its stopped target,
where $\bar\phi$ is an exponential moving average of $\phi$.
The local state is $s_t^i=[h_t^i;\operatorname{vec}(z_t^i)]$.
Unless stated otherwise, per-transition losses are averaged over their
valid entries; invalid entries contribute zero, and an empty set of
valid entries contributes zero. For the local objectives, validity is
given by the replay presence and learning-activity masks, not
necessarily physical controllability.

The two local JEPA objectives are
\begin{align}
    \mathcal L_{\mathrm{post}}
    &=
    \mathbb E_{\mathrm{valid}}\!\bigl[
        d_{\cos}(P_{\mathrm{post}}(s_t^i),\bar e_t^i)
    \bigr], \\
    \mathcal L_{\mathrm{dyn}}
    &=
    \mathbb E_{\mathrm{valid}}\!\bigl[
        d_{\cos}(P_{\mathrm{dyn}}(h_t^i),\bar e_t^i)
    \bigr],
\end{align}
where $d_{\cos}(x,y)=1-\operatorname{sim}_{\cos}(x,y)$.
The posterior and prior are
$q_t^i=q_\theta(\cdot\mid h_t^i,e_t^i)$ and
$p_t^i=p_\theta(\cdot\mid h_t^i)$, respectively. Both use categorical
probabilities with uniform mixing coefficient $0.01$. The two directed
regularizers are
\begin{align}
    \mathcal L_{\mathrm{dyn-reg}}
    &=
    \mathbb E_{\mathrm{valid}}\!\left[
        \max\!\left\{
            \tau,
            D_{\mathrm{KL}}
            \bigl(\operatorname{sg}(q_t^i)\|p_t^i\bigr)
        \right\}
    \right], \\
    \mathcal L_{\mathrm{rep-reg}}
    &=
    \mathbb E_{\mathrm{valid}}\!\left[
        \max\!\left\{
            \tau,
            D_{\mathrm{KL}}
            \bigl(q_t^i\|\operatorname{sg}(p_t^i)\bigr)
        \right\}
    \right],
\end{align}
with free-nat threshold $\tau=1$. KL divergences sum over the
categorical variables before applying this threshold.

The complete local objective is
\begin{equation}
\begin{aligned}
    \mathcal L_{\mathrm{local}}
    ={}&
    2\mathcal L_{\mathrm{post}}
    +2\mathcal L_{\mathrm{dyn}}
    +0.05\mathcal L_{\mathrm{SIGReg}} \\
    &+
    \mathcal L_{\mathrm{mask}}^{\mathrm{local}}
    +\mathcal L_{\mathrm{dyn-reg}}
    +0.1\mathcal L_{\mathrm{rep-reg}}.
\end{aligned}
\label{eq:local-full-objective}
\end{equation}
There are no local reward or continuation losses: these outcomes are
predicted by the joint model. The vector-observation configuration
contains neither an observation-reconstruction objective nor a spatial
prediction objective.

\paragraph{Representation regularization.}
SIGReg is applied to online embeddings, grouping samples by agent slot across batch and time. For each nonempty group, it compares the empirical characteristic functions of 256 random unit-vector projections with the standard Gaussian characteristic function.
The comparison uses 17 equally spaced points in $[0,3]$ and Gaussian-weighted quadrature. The projected discrepancy is multiplied by the
group's valid sample count before averaging over nonempty groups.
This sample-count scaling is part of the objective.

\paragraph{Action-availability supervision.}
The local availability head predicts a binary vector from $s_t^i$.
For each vector, binary cross-entropy is averaged separately over legal and illegal actions, then averaged equally over the nonempty classes.
The joint availability head uses the same balanced reduction.

\subsection{One-Step Joint Prediction and Posterior Alignment}
\label{app:joint-objective}

The joint model receives synchronized local states, the joint action,
roster presence, and controllability. Agent self-attention first mixes
information within a transition; a causal Transformer then maintains
temporal context. Writing its cache explicitly as $k_t$, we obtain
\begin{align}
    (k_{t+1},u_t^{1:N})
    &=
    J_\omega\!\left(
        k_t,
        \operatorname{sg}(s_t^{1:N}),
        a_t^{1:N},
        m_{\mathrm{pres},t}^{1:N},
        m_{\mathrm{ctrl},t}^{1:N}
    \right), \\
    \hat e_{t+1}^i
    &=P_e(u_t^i).
\end{align}
No fixed agent-identity or agent-position embeddings are added.
The detached local-state input specifies the baseline gradient boundary.

The embedding and interface objectives are
\begin{align}
    \mathcal L_{\mathrm{emb}}
    &=
    \mathbb E_{\mathrm{valid}}\!\bigl[
        d_{\cos}(\hat e_{t+1}^i,\bar e_{t+1}^i)
    \bigr], \\
    \mathcal L_{\mathrm{int}}
    &=
    \mathbb E_{\mathrm{valid}}\!\bigl[
        \operatorname{SmoothL1}
        (\hat e_{t+1}^i,\operatorname{sg}(e_{t+1}^i))
    \bigr].
\end{align}
Smooth-$L_1$ is averaged over embedding coordinates and uses a
quadratic-to-linear transition at absolute error one. It penalizes
pointwise discrepancy with the online encoder output; it does not
establish equality of the real and imagined embedding distributions.

We additionally align the posterior distributions induced by real and
predicted embeddings while holding the factual history fixed:
\begin{equation}
\begin{aligned}
    \mathcal L_{\mathrm{align}}
    =
    \mathbb E_{\mathrm{valid}}\!\Bigl[
        D_{\mathrm{KL}}\!\bigl(
        &\operatorname{sg}
        (q_\theta(\cdot\mid h_{t+1}^i,e_{t+1}^i))
        \,\| \\
        &q_{\theta^-}
        (\cdot\mid\operatorname{sg}(h_{t+1}^i),\hat e_{t+1}^i)
        \bigr)
    \Bigr],
\end{aligned}
\label{eq:one-step-posterior-alignment}
\end{equation}
where $\theta^-$ denotes a frozen copy of the current local parameters,
not a separate learned posterior. The KL uses the same uniform-mixed
categorical distributions as execution. Gradients pass through the
predicted embedding to the joint model, but not to the factual history,
target posterior, or local posterior parameters along this path.

The one-step joint objective is
\begin{equation}
\begin{aligned}
    \mathcal L_{\mathrm{joint}}
    ={}&
    2\mathcal L_{\mathrm{emb}}
    +\mathcal L_{\mathrm{int}}
    +0.05\mathcal L_{\mathrm{align}} \\
    &+
    \mathcal L_r^{\mathrm{joint}}
    +\mathcal L_c^{\mathrm{joint}}
    +\mathcal L_{\mathrm{mask}}^{\mathrm{joint}}
    +\mathcal L_{\mathrm{alive}}.
\end{aligned}
\label{eq:joint-full-objective}
\end{equation}
Reward prediction uses a symlog two-hot categorical output. Continuation
and controllability use binary outputs. With discounted continuation,
the continuation target is $\gamma(1-d_{t+1}^i)$, where $d$ denotes the
replay terminal flag and $\gamma=1-1/333$.

Embedding, interface, alignment, and next-mask losses require a
controllable source, a present target, and no reset crossing. We still predict when death occurs after the transition. Reward and
continuation losses also supervise dead but present roster slots.
Learning when agents are alive is supervised only on unmasked slots. These masks are normalized separately.

\subsection{Direct Multi-Step JEPA and Action Discrimination}
\label{app:multistep}

Direct prediction heads are supervised at
$\mathcal H_{\mathrm{direct}}=\{1,2,4,8\}$.
The factual interaction feature $u_t^i$ already contains the current
joint action $a_t^{1:N}$. For $h>1$, the corresponding head also receives
the focal agent's recorded action tail:
\begin{equation}
    \hat e_{t+h\mid t}^i
    =
    P_h(u_t^i,a_{t+1:t+h-1}^i).
\end{equation}
The tail is empty for $h=1$. Later teammate actions are not supplied by
a teammate-plan model. These predictions are evaluated directly from
factual roots; they are neither chained between horizons nor used to
construct PPO states.

For ordered horizons $(h_0,h_1,h_2,h_3)=(1,2,4,8)$, define
\begin{equation}
    \alpha_{h_k}
    =
    \frac{\rho^k}{\sum_{j=0}^{3}\rho^j},
    \qquad
    \rho=0.75.
\end{equation}
The direct loss is
\begin{equation}
    \mathcal L_{\mathrm{MS}}
    =
    \sum_{h\in\mathcal H_{\mathrm{direct}}}
    \alpha_h
    \mathbb E_{\mathrm{valid},h}\!\bigl[
        d_{\cos}(\hat e_{t+h\mid t}^i,\bar e_{t+h}^i)
    \bigr].
    \label{eq:direct-multistep}
\end{equation}
The available root indices leave room for the maximum horizon in the
sampled suffix. Each horizon additionally requires a legal factual
prefix, present and controllable action sources, and no reset crossing.
Death exactly at the target is retained.

For each $h>1$, action discrimination replaces the final focal-agent
action $a_{t+h-1}^i$ with every distinct alternative legal at that
factual decision. The joint root and earlier focal actions remain
unchanged. Each alternative incurs
\begin{equation}
\begin{aligned}
    \ell_{\mathrm{AD},h}^i
    =
    \Bigl[
    &m-
    \operatorname{sim}_{\cos}
    (\hat e_{t+h\mid t}^i,\bar e_{t+h}^i) \\
    &+
    \operatorname{sim}_{\cos}
    (\hat e_{t+h\mid t}^{i,\mathrm{alt}},\bar e_{t+h}^i)
    \Bigr]_+,
    \qquad m=0.1.
\end{aligned}
\label{eq:all-legal-margin}
\end{equation}
We first average over legal alternatives at each root and then over
valid roots. Horizon weights are renormalized over $\{2,4,8\}$ for the
action loss. The resulting auxiliary is
$2\mathcal L_{\mathrm{MS}}+0.1\mathcal L_{\mathrm{AD}}$.
The alternatives are discriminative negatives, not observed
counterfactual targets; different legal actions need not always yield
different observable outcomes.

\subsection{Self-Fed Recurrent Supervision}
\label{app:self-fed}

In addition to direct prediction, we train recursive joint predictions
under recorded joint actions. Starting from a factual root, the joint
model predicts the next observation embedding, the local posterior
constructs the next latent state, and that predicted state becomes the
input to the following transition. Real future observations are used
as targets, not injected into the predicted trajectory.

We sample up to eight eligible synchronized roots without replacement
from the world-model batch. A root requires two valid subsequent
transitions; later endpoint validity is assessed separately. The rollout extends for five transitions, with supervision at
$\mathcal H_{\mathrm{SF}}=\{2,4,5\}$.
Terminal arrivals are retained, while reset crossings, disappearing
slots, and out-of-range endpoints are masked. Individual death does
not itself define an episode boundary.

The joint model and reconstructed joint history use inference mode.
Local transition parameters are frozen. Gradients through the joint
temporal cache and the permitted local input paths are truncated by
detaching recurrent state at the start of every two-step chunk.
The joint predictor still detaches its local-state input at each step;
this is not unrestricted end-to-end backpropagation through the full
five-step simulator. The trajectory-alignment term retains the input
Jacobian through the frozen local transition within each chunk.

Unlike one-step alignment, trajectory alignment compares factual and
recursively generated histories:
\begin{equation}
\begin{aligned}
    \mathcal L_{\mathrm{traj},h}
    =
    \mathbb E_{\mathrm{valid},h}\!\Bigl[
        D_{\mathrm{KL}}\!\bigl(
        &\operatorname{sg}
        (q_\theta(\cdot\mid h_{t+h}^i,e_{t+h}^i))
        \,\| \\
        &q_{\theta^-}
        (\cdot\mid\hat h_{t+h}^i,\hat e_{t+h}^i)
        \bigr)
    \Bigr].
\end{aligned}
\label{eq:trajectory-posterior-alignment}
\end{equation}
The total recurrent auxiliary is
\begin{equation}
\begin{aligned}
    \mathcal L_{\mathrm{SF}}
    =
    \frac{0.1}{|\mathcal H_{\mathrm{SF}}|}
    \sum_{h\in\mathcal H_{\mathrm{SF}}}
    \Bigl(
    &2\mathcal L_{\mathrm{emb},h}
    +\mathcal L_{\mathrm{int},h}
    +\mathcal L_{r,h}^{\mathrm{joint}} \\
    &+\mathcal L_{c,h}^{\mathrm{joint}}
    +\mathcal L_{\mathrm{mask},h}^{\mathrm{joint}}
    +\mathcal L_{\mathrm{alive},h}
    +0.1\mathcal L_{\mathrm{traj},h}
    \Bigr).
\end{aligned}
\label{eq:self-fed-objective}
\end{equation}
Each endpoint loss is normalized over its valid roots and agents;
endpoints receive equal weight. The trajectory-KL coefficient is
therefore $0.01$ relative to the endpoint-averaged trajectory loss.
Shared reward and continuation losses include dead but present slots;
the other endpoint losses require a controllable root.

This supervised rollout uses the joint availability head with
thresholded masks. Recorded actions continue to drive the rollout even
when excluded by its predicted mask. This differs from the local-head,
Bernoulli-sampled masks used for PPO imagination below.

\subsection{Joint Imagination and Decentralized Execution}
\label{app:joint-imagination-details}

After each world-model update, we reconstruct the independent behavior
batch with the updated model. Its 64 suffix positions supply candidate
imagination roots. The episode-final roots are excluded by validity masks.
Each valid root produces $H=5$ imagined transitions.

At every transition, each agent samples an action from the shared actor
using only its own local state and current action mask. The local
history advances using that agent's latent and action. In parallel, the
joint predictor consumes the synchronized states, joint action, and
joint temporal cache to predict the next local embedding. The next
stochastic state is then sampled as
\begin{equation}
    \hat z_{t+1}^i
    \sim
    q_\theta(\cdot\mid\hat h_{t+1}^i,\hat e_{t+1}^i).
\end{equation}
The local prior remains a training regularizer, but it is not the
sampling distribution used for these joint imagined transitions.

\paragraph{Imagined action availability.}
Root masks come from replay. At subsequent states, the local
availability head predicts probabilities from the completed local
state. Each action's availability is sampled as a Bernoulli variable.
An empty predicted mask falls back to the no-op action, and
non-controllable agents receive no-op-only support. The effective masks
actually used to sample actions are stored with the imagined batch and
are not resampled during PPO updates.

\paragraph{SMAC roster and team outcomes.}
The fixed roster is retained after individual deaths. Imagined
controllability is monotonic: an agent remains controllable only if it
was controllable previously and the predicted next controllability
probability is at least $0.5$. Joint reward and continuation predictions
are averaged over present slots and broadcast back to those slots.
The all-dead team is absorbing: the transition into this condition
retains its final reward, while continuation and later rewards are zero.
These are the shared-reward, fixed-roster rules of the SMAC adapter,
not assumptions about all benchmark environments.

\paragraph{Execution.}
Real interaction uses the observation encoder, local history model, local posterior, and actor, together with the environment-provided
legal-action mask. Parameters are shared, but each agent maintains its
own history. Neither the joint predictor nor the centralized critic is
used during execution. There is no teammate-belief network or residual
belief adapter in this configuration.

\subsection{PPO and Replay-Value Learning}
\label{app:ppo}

One imagined batch is generated before actor--critic optimization.
Its local features, centralized critic context, actions, effective
masks, old policy logits, advantages, returns, and trajectory weights
remain fixed throughout the update. The world model receives no PPO
gradients.

Let $\bar V_t^i$ be the target critic prediction and let $b_t^i$
indicate value-valid states. Define the effective discounted
continuation $\tilde c_{t+1}^i=\hat c_{t+1}^i b_{t+1}^i$.
Generalized advantages and return targets are
\begin{align}
    \delta_t^i
    &=
    \hat r_{t+1}^i
    +\tilde c_{t+1}^i\bar V_{t+1}^i
    -\bar V_t^i, \\
    A_t^i
    &=
    b_t^i\bigl(
        \delta_t^i
        +\lambda\tilde c_{t+1}^i A_{t+1}^i
    \bigr),
    \qquad A_H^i=0, \\
    R_t^i
    &=
    A_t^i+\bar V_t^i,
    \qquad \lambda=0.95.
\end{align}
The continuation already includes discounting; it is not multiplied by
$\gamma$ a second time. Critic-valid states include dead but present
SMAC agents, whereas actor-valid decisions additionally require
controllability. An individual death therefore removes subsequent
actor decisions without discarding the earlier shared-team return.

Losses use cumulative continuation weights,
$w_t^i=b_t^i\prod_{j=1}^{t}\tilde c_j^i$, with an empty product of one.
Actor advantages are normalized once using the weighted mean and
variance over valid decisions in the frozen batch. Writing the
normalized advantage as $\widetilde A_t^i$ and the stored mask as $M_t^i$,
the probability ratio is
\begin{equation}
    r_t^i(\psi)
    =
    \frac{\pi_\psi(a_t^i\mid s_t^i,M_t^i)}
         {\pi_{\mathrm{old}}(a_t^i\mid s_t^i,M_t^i)}.
\end{equation}
For numerical stability, its log is clipped to $[-20,20]$ before
exponentiation. The actor minimizes
\begin{equation}
\begin{aligned}
    \mathcal L_{\mathrm{actor}}
    =
    -\mathbb E_{\mathrm{actor},w}\Bigl[
        \min\bigl(
        &r_t^i\widetilde A_t^i,
        \operatorname{clip}(r_t^i,1-\epsilon,1+\epsilon)
        \widetilde A_t^i
        \bigr) \\
        &+\beta_{\mathrm{ent}}
        \mathcal H\bigl(
            \pi_\psi(\cdot\mid s_t^i,M_t^i)
        \bigr)
    \Bigr],
\end{aligned}
\end{equation}
where $\mathbb E_{\mathrm{actor},w}$ is the normalized weighted mean
over actor-valid decisions, $\epsilon=0.2$, and
$\beta_{\mathrm{ent}}=0.003$. Entropy is not divided by the size of the
action support.

The centralized critic predicts a symlog two-hot value distribution and
is trained with that distribution's target loss, rather than a clipped
mean-squared-error value objective:
\begin{equation}
    \mathcal L_{\mathrm{critic}}
    =
    \mathbb E_{\mathrm{value},w}\bigl[
        \ell_{\mathrm{twohot}}(V_\xi,\operatorname{sg}(R))
    \bigr]
    +0.3\mathcal L_{\mathrm{replay-value}}.
\end{equation}
The replay-value term uses real rewards and current imagined
root-return bootstraps to construct $\lambda$-returns with
$\lambda_{\mathrm{replay}}=0.95$. It trains only the critic; recorded
replay actions are not PPO actor samples. Terminal arrivals retain
their rewards and stop bootstrapping, and traces never cross resets.
When the adapter distinguishes nonterminal truncation, the final real
observation supplies its bootstrap. The supplied SMAC adapter uses the
same upstream termination flag for episode-final and terminal markers,
so that distinct truncation case is not exercised by this adapter.

Five actor steps and five critic steps reuse the same full imagined
batch, with actor and critic updates interleaved. No new trajectories
or action masks are sampled between these steps. The target critic is
copied from the updated critic after each active learner call; the
additional slow-critic regularizer is disabled.

\subsection{Optimizer Ownership and Update Order}
\label{app:optimizer-ownership}

Parameters are partitioned into four disjoint groups: the local world
model, joint world model, actor, and centralized critic. The model
objective is
\begin{equation}
    \mathcal L_{\mathrm{WM}}
    =
    \mathcal L_{\mathrm{local}}
    +\mathcal L_{\mathrm{joint}}
    +2\mathcal L_{\mathrm{MS}}
    +0.1\mathcal L_{\mathrm{AD}}
    +\mathcal L_{\mathrm{SF}}.
\end{equation}
The local group contains the encoder, local dynamics and posterior,
local JEPA predictors, and local availability head. The joint group contains the interaction and temporal predictor, joint output heads,
and direct multi-step heads. In the baseline, joint losses do not
update the local-world parameters. Frozen local modules can still
transmit gradients with respect to their inputs on the alignment and
self-fed paths described above.

A learner call first differentiates $\mathcal L_{\mathrm{WM}}$ and
updates the two model groups, leaving actor and critic optimizer states
untouched. It then reconstructs the behavior view with the updated
model, generates and freezes an imagined batch, and performs the
separate PPO actor and critic steps. This is not a single combined
world-model--actor--critic backward pass.

All groups use adaptive gradient clipping followed by RMS
preconditioning and momentum. The local and joint world-model learning
rates are $10^{-4}$; actor and critic learning rates are
$3\times10^{-5}$. The clipping coefficient is $0.3$, momentum
coefficients are $(0.9,0.999)$, optimizer $\epsilon$ is $10^{-20}$,
and weight decay and learning-rate warm-up are disabled in this
configuration. The target encoder is updated after a finite model
update using
\begin{equation}
    \bar\phi
    \leftarrow
    0.99\bar\phi+0.01\phi.
\end{equation}

\subsection{Baseline Hyperparameters}
\label{app:baseline-configuration}

We benchmark DMAWM, MAPPO, QMIX, and MAT using
implementations released by their respective authors. The tables below
report the effective settings used by our benchmark launchers, verified
against the released implementations and saved run configurations.

Unless stated otherwise, all baselines use StarCraft~II~4.10.0, AI difficulty 7, and \texttt{step\_mul}=8.
Training budgets count real environment transitions across collectors;
evaluation interactions, imagined transitions, and individual-agent
actions do not contribute to this count.

\newpage
\subsubsection{MA-JEPA Hyperparameters}
\label{app:configuration}
\label{app:majepa-hyperparameters}

We use a shared configuration across all SMAC tasks, changing only the task,
number of agents, training seed, and interaction budget. 

\begin{table}[!htbp]
    \centering
    \caption{MA-JEPA hyperparameters used in our SMAC experiments.}
    \label{tab:majepa-hyperparameters}
    \small
    \renewcommand{\arraystretch}{1.08}
    \begin{tabular}{@{}p{0.58\linewidth}p{0.36\linewidth}@{}}
        \toprule
        \textbf{Hyperparameter} & \textbf{Value used} \\
        \midrule
        \multicolumn{2}{@{}l}{\textit{World model}} \\
        Observation encoder layers / width & 3 / 1024 \\
        Deterministic state dimension & 4096 \\
        Categorical latent & $32\times64$ \\
        Latent uniform mixture & 0.01 \\
        Local Transformer layers / width / heads & 2 / 512 / 8 \\
        Local / joint context length & 64 / 16 \\
        Joint agent-interaction / temporal layers & 2 / 12 \\
        Joint Transformer width / heads & 256 / 4 \\
        Joint dropout & 0.1 \\
        World-model learning rate & $10^{-4}$ \\
        Adaptive gradient clipping & 0.3 \\
        Target-encoder EMA rate & 0.01 \\
        SIGReg coefficient & 0.05 \\
        Direct JEPA prediction horizons & $\{1,2,4,8\}$ \\
        Action-margin / loss coefficient & 0.1 / 0.1 \\
        Self-fed horizons / BPTT length & $\{2,4,5\}$ / 2 \\
        Outcome / factual JEPA input-gradient scales & 1.0 / 0.1 \\
        \midrule
        \multicolumn{2}{@{}l}{\textit{Policy learning}} \\
        Actor hidden layers / width & 3 / 512 \\
        Central critic layers / width & 2 / 256 \\
        Actor / critic learning rate & $3\times10^{-5}$ / $3\times10^{-5}$ \\
        Actor / critic PPO epochs & 5 / 5 \\
        PPO clipping & 0.2 \\
        Entropy coefficient & 0.003 \\
        Discount / return-trace $\lambda$ & $1-1/333$ / 0.95 \\
        Imagination horizon & 5 \\
        Auxiliary replay-value loss coefficient & 0.3 \\
        Imagined action mask & Local; Bernoulli sampling \\
        \bottomrule
    \end{tabular}
\end{table}

\newpage
\subsubsection{DMAWM}
\label{app:dmawm-hyperparameters}

DMAWM follows \citet{Xue2026} and the authors' released implementation.\footnote{
\url{https://github.com/DiXue98/DMAWM/tree/2acaaeb82805b55d275e3ce08ac8f713ec9afbb2}}

\begin{table}[!htbp]
    \centering
    \caption{DMAWM hyperparameters used in our benchmark.}
    \label{tab:dmawm-hyperparameters}
    \small
    \renewcommand{\arraystretch}{1.08}
    \begin{tabular}{@{}p{0.58\linewidth}p{0.36\linewidth}@{}}
        \toprule
        \textbf{Hyperparameter} & \textbf{Value used} \\
        \midrule
        \multicolumn{2}{@{}l}{\textit{Reinforcement learning}} \\
        Optimizer
            & Adam \\
        Actor / critic learning rate
            & $3\times10^{-5}$ / $3\times10^{-5}$ \\
        Actor / critic gradient clipping
            & 100 / 100 \\
        Entropy coefficient
            & 0.01 \\
        PPO epochs / clipping
            & 5 / 0.2 \\
        Discount $\gamma$ / GAE $\lambda$
            & 0.99 / 0.95 \\
        Replay-value loss coefficient
            & 1.0 \\
        Importance-weight cap
            & 4 \\
        Actor / critic sharing
            & True / True \\
        \midrule
        \multicolumn{2}{@{}l}{\textit{World model}} \\
        World-model learning rate
            & $10^{-4}$ \\
        World-model gradient clipping
            & \textbf{1000} \\
        Batch size / sequence length
            & 16 / 64 \\
        Replay burn-in / imagination horizon
            & 4 / 4 \\
        Replay capacity / prefill
            & $250{,}000$ / 5000 \\
        Training ratio
            & 128 \\
        Dynamics / representation KL scales
            & 1.0 / 0.1 \\
        KL free bits
            & 1.0 \\
        Deterministic latent width
            & 512 \\
        Categorical latent
            & $32\times32$ \\
        Latent uniform mixture
            & 0.01 \\
        Transition / global Transformer layers
            & 3 / 3 \\
        Transformer heads
            & 8 \\
        Observation decoder hidden width / layers
            & 1024 / 2 \\
        \midrule
        Collection environments
            & 16 \\
        Actor / critic MLP width / layers
            & 256 / 2 \\
        Critic Transformer layers / heads
            & 1 / 8 \\
        \bottomrule
    \end{tabular}
\end{table}

\newpage
\subsubsection{MAPPO and MAT}
\label{app:mappo-mat-hyperparameters}

MAPPO uses the official implementation of \citet{yu2022mappo}.\footnote{
\url{https://github.com/marlbenchmark/on-policy/tree/de66d7a4b23fac2513f56f96f73b3f5cb96695ac}}
MAT uses the official implementation of \citet{wen2022mat}.\footnote{
\url{https://github.com/PKU-MARL/Multi-Agent-Transformer/tree/be3ff49c8264d454c1fe2c41582aa2bfc98498c8}}

\begin{table}[!htbp]
    \centering
    \caption{MAPPO and MAT hyperparameters used in our benchmark.}
    \label{tab:mappo-mat-hyperparameters}
    \small
    \renewcommand{\arraystretch}{1.08}
    \begin{tabular}{@{}p{0.47\linewidth}p{0.22\linewidth}p{0.23\linewidth}@{}}
        \toprule
        \textbf{Hyperparameter}
        & \textbf{MAPPO}
        & \textbf{MAT} \\
        \midrule
        Policy
            & Recurrent or FF
            & Transformer \\
        Optimizer
            & Adam
            & Adam \\
        Actor / critic learning rate
            & $5\times10^{-4}$ each
            & Joint $5\times10^{-4}$ \\
        Rollout length per collector
            & 400
            & 100 \\
        Parallel collection environments
            & 8
            & 32 \\
        Transitions per rollout batch
            & 3200
            & 3200 \\
        Entropy coefficient
            & 0.01
            & 0.01 \\
        Discount $\gamma$
            & 0.99
            & 0.99 \\
        GAE $\lambda$
            & 0.95
            & 0.95 \\
        PPO epochs
            & 15 or 5
            & 15 \\
        PPO clipping
            & 0.2 or 0.05
            & 0.05 \\
        Minibatches per PPO epoch
            & 1
            & 1 \\
        Maximum gradient norm
            & 10
            & 10 \\
        MLP / recurrent hidden width
            & 64
            & --- \\
        Recurrent layers
            & 1
            & --- \\
        Transformer blocks / heads
            & ---
            & 1 / 1 \\
        Transformer embedding width
            & ---
            & 64 \\
        Value normalization / PopArt
            & True / False
            & True / False \\
        Policy / value active masks
            & True / False
            & True / False \\
        \bottomrule
    \end{tabular}
\end{table}

\subsubsection{QMIX}
\label{app:qmix-hyperparameters}

We evaluate QMIX \citep{rashid2018qmix} using the optimized PyMARL2
implementation.\footnote{
\url{https://github.com/hijkzzz/pymarl2/tree/8ccac7c5aa134422a2e3009be735d23cdf8ce2f8}}
The settings combine PyMARL2's QMIX, default, and SMAC configurations.

\begin{table}[!htbp]
    \centering
    \caption{QMIX/PyMARL2 hyperparameters used in our benchmark.}
    \label{tab:qmix-hyperparameters}
    \small
    \renewcommand{\arraystretch}{1.08}
    \begin{tabular}{@{}p{0.58\linewidth}p{0.36\linewidth}@{}}
        \toprule
        \textbf{Hyperparameter} & \textbf{Value used} \\
        \midrule
        Agent architecture / hidden width
            & Recurrent / 64 \\
        Optimizer / learning rate
            & Adam / $10^{-3}$ \\
        Discount $\gamma$ / TD $\lambda$
            & 0.99 / 0.6 \\
        Target-network update interval
            & 200 episodes \\
        Maximum gradient norm
            & 10 \\
        Training batch size
            & 128 episodes \\
        Replay capacity
            & 5000 episodes \\
        Parallel collection environments
            & 8 \\
        Exploration $\epsilon$
            & $1.0\rightarrow0.05$ \\
        Linear epsilon annealing duration
            & 100K environment steps \\
        Mixer embedding / hypernetwork width
            & 32 / 64 \\
        Agent-ID / previous-action inputs
            & True / True \\
        Layer normalization
            & False \\
        Orthogonal initialization
            & False \\
        Prioritized replay
            & False \\
        \bottomrule
    \end{tabular}
\end{table}

\section{Consistency of Learned Joint Successor Law with Fixed Representations}
\label{app:additional-theory}

\paragraph{Consistency of the joint successor law.} Fix \method{}'s parameters and collection policy. Consider independent and identically distributed episodes of $T<\infty$ steps. We use $\mathbf{x}$ to refer to the vector consisting of some agent-specific variable $x_i$ collected across all agents $i\in N$. Let $\mathbf z_t$ denote
the joint latent state, $\mathbf{a}_t\in\mathcal{A}$ the  action, and $\mathbf{m}_t$ the joint legal-action mask. For a fixed history-window length
$w\geq1$, define the ``padded context''
$\mathbf c_t=(t,\mathbf z_{t-w+1:t},\mathbf a_{t-w:t-1},
\mathbf m_{t-w+1:t})$
and the complete successor state
$\mathbf y_{t+1}=(\mathbf z_{t+1},\tilde r_{t+1},
e_{t+1},\mathbf m_{t+1})\in\mathcal Y$.
Here $\tilde r$ is a quantized reward and $e\in\{0,1\}$ indicates termination.
All alphabets are finite. Let $P(\cdot\mid\mathbf c,\mathbf a)$ denote the
conditional distribution of the complete successor under the fixed parameters
and collection policy. We suppress the conditioning arguments in the bounds below.
\begin{theoremE}[Joint-law consistency][end, restate, text link section]
\label{thm:discrete-tv}
From $n$ sampled episodes, let $n(\mathbf c,\mathbf a)$ count occurrences of
context--action pair $(\mathbf c,\mathbf a)$. At each observed pair, let
$\widehat P_n(\cdot\mid\mathbf c,\mathbf a)$ be the empirical distribution
of its complete successors. At an unobserved pair, define $\widehat P_n$ to
be any fixed probability distribution on $\mathcal Y$.

As $n\to\infty$, $\widehat P_n$ converges to $P$ almost surely in total
variation, uniformly over all context--action pairs with positive probability
of occurring under the fixed collection policy. For fixed $n$, $n_0\geq1$,
and $0<\delta<1$, the following bound holds with probability at least
$1-\delta$, simultaneously for all pairs with $n(\mathbf c,\mathbf a)\geq n_0$.
\[
 \operatorname{TV}(\widehat P_n,P)
 \leq\min\Bigl\{1,\sqrt{\tfrac{
 |\mathcal Y|\log2+\log(2|\mathcal C||\mathcal A|/\delta)
 }{2n_0}}\Bigr\}.
\]
Conditional on observing a given pair exactly $s\geq1$ times, its expected
total-variation error satisfies
\[
 \mathbb E\bigl[\operatorname{TV}(\widehat P_n,P)
 \mid n(\mathbf c,\mathbf a)=s\bigr]
 \leq\tfrac12\sqrt{(|\mathcal Y|-1)/s}.
\]
At each context--action pair with positive occurrence probability, consider
an unrestricted autoregressive model of the complete successor $\mathbf y$.
The cross-entropy between the learned and true population distribution over next latents is uniquely minimized by the
joint distribution $P(\cdot\mid\mathbf c,\mathbf a)$.
\end{theoremE}
\begin{proofE}
Condition on the episodes containing a fixed positive-probability pair
$(\mathbf c,\mathbf a)$. Our setting ensures one outcome per
selected episode, so these outcomes are iid from $P$.
Hoeffding's inequality over the $2^{|\mathcal Y|}$ outcome subsets,
averaged over counts $s\geq n_0$, gives, for $\eta>0$,
\[
 \Pr\bigl(\TV(\widehat P_n,P)>\eta,\ n(\mathbf c,\mathbf a)\geq n_0\bigr)
 \leq2^{|\mathcal Y|+1}e^{-2n_0\eta^2}.
\]
A union bound over pairs gives the probability bound. For
$p_{\mathbf y}=P(\mathbf y\mid\mathbf c,\mathbf a)$, binomial variance and
Cauchy--Schwarz give
\[
 \E[\TV(\widehat P_n,P)\mid s]
 \leq\frac1{2\sqrt{s}}\sum_{\mathbf y}\sqrt{p_{\mathbf y}(1-p_{\mathbf y})}
 \leq\frac12\sqrt{\frac{|\mathcal Y|-1}{s}}.
\]
For positive-probability pairs, the strong law gives convergence of the
empirical count ratios. Finitely many pairs and outcomes then give
uniform almost-sure convergence.

For the autoregressive claim, fix such a pair and order the coordinates of
the complete outcome $\mathbf y$. Every joint law admits a factorization
$\widetilde q(\mathbf y)=\prod_j\widetilde q_j(y_j\mid y_{<j})$.
Its teacher-forced cross-entropy decomposes into entropy and
Kullback--Leibler divergence:
\[
 -\E_P\sum_j\log\widetilde q_j(y_j\mid y_{<j})
 =H(P)+D_{\mathrm{KL}}(P\Vert\widetilde q),
\]
so its unique minimizing joint law is $P$.
\end{proofE}

\section{Proofs of Theoretical Results}
\printProofs
\end{document}